%% file: aaai20.tex
\def\year{2020}\relax
\documentclass[letterpaper]{article} 
\usepackage{aaai20}  
\usepackage{times}  
\usepackage{helvet} 
\usepackage{courier}  
\usepackage[hyphens]{url}  
\usepackage{graphicx} 
\usepackage{graphicx}  
\usepackage[square]{natbib}
\usepackage[utf8]{inputenc} 
\usepackage[T1]{fontenc}    
\usepackage{hyperref}

\usepackage{booktabs}       
\usepackage{amsfonts}       
\usepackage{nicefrac}       
\usepackage{microtype}      

\usepackage{epsfig}
\usepackage{amsmath}
\usepackage{amssymb}
\usepackage{adjustbox}

\usepackage{amsthm}  
\usepackage{wasysym}
\usepackage[ruled,vlined,algo2e]{algorithm2e}
\providecommand{\SetAlgoLined}{\SetLine}
\usepackage{color}
\usepackage{dsfont}	
\usepackage{wrapfig}
\usepackage{footnote}
\usepackage{epstopdf}
\usepackage{enumitem}
\usepackage{subfigure}
\usepackage{caption}
\usepackage{comment}
\usepackage{multirow}

\usepackage{algorithm}
\usepackage{algorithmic}

\def\eg{\emph{e.g. }}
\def\ie{\emph{i.e. }}

\def\vs{\emph{vs. }}
\def\wrt{\emph{w.r.t. }}
\def\etal{\emph{et al. }}

\def\ride{{\sc Ride}}

\usepackage{pifont}
\newcommand{\cmark}{\ding{51}}%

\DeclareMathOperator*{\argmin}{arg\,min}

\usepackage{mathtools}

\DeclarePairedDelimiter\floor{\lfloor}{\rfloor}

\makeatletter
\newcommand*{\rom}[1]{\expandafter\@slowromancap\romannumeral #1@}
\makeatother
\makeatletter
\newcommand\footnoteref[1]{\protected@xdef\@thefnmark{\ref{#1}}\@footnotemark}
\makeatother
\newcommand{\bfsection}[1]{\vspace*{0.1cm}\noindent\textbf{#1.}}

\newtheorem{thm}{Theorem}

\begin{document}

\title{White-Box Adversarial Defense via Self-Supervised Data Estimation}
\author{Zudi Lin\textsuperscript{\rm 1}\thanks{Work was done during an internship at MERL.}, Hanspeter Pfister\textsuperscript{\rm 1} and Ziming Zhang\textsuperscript{\rm 2}\thanks{Corresponding author.}\\
\textsuperscript{\rm 1}Harvard University, \textsuperscript{\rm 2}Mitsubishi Electric Research Laboratories (MERL)\\
\texttt{\{linzudi,pfister\}@g.harvard.edu,zzhang@merl.com}
}
\maketitle
\begin{abstract}

In this paper, we study the problem of how to defend classifiers against adversarial attacks that fool the classifiers using subtly modified input data. In contrast to previous works, here we focus on the {\em white-box} adversarial defense where the attackers are granted full access to not only the classifiers but also defenders to produce as strong attacks as possible. In such a context we propose viewing a defender as a {\em functional}, a higher-order function that takes functions as its argument to represent a {\em function space}, rather than fixed functions conventionally. From this perspective, a defender should be realized and optimized {\em individually} for each adversarial input. To this end, we propose {\ride}, an efficient and provably convergent self-supervised learning algorithm for individual data estimation to protect the predictions from adversarial attacks.  We demonstrate the significant improvement of adversarial defense performance on image recognition, \eg, {98\%, 76\%, 43\%} test accuracy on MNIST, CIFAR-10, and ImageNet datasets respectively under the state-of-the-art BPDA attacker. 
\end{abstract}

\input{introduction.tex}
\input{related.tex}
\input{motivation.tex}
\input{method.tex}
\input{experiment.tex}

\section{Conclusion}


In this paper, we propose a novel functional-based adversarial defender, {\ride}, against white-box adversarial attacks without any modification of well-trained classifiers. Our defender utilizes an iterative self-supervised optimization algorithm to estimate the clean data for each individual adversarial input. For future work, we will focus on improving the optimizing efficiency of {\ride} and target for real-time applications.

\newpage
{\small
\bibliographystyle{aaai}
\bibliography{egbib}
}  

\input{appendix.tex}

\end{document}

%% file: introduction.tex
\section{Introduction}

Recent studies have demonstrated that as classifiers, deep neural networks (\eg, CNNs) are quite vulnerable to adversarial attacks that only add quasi-imperceptible perturbations to the input data but completely change the predictions of the classifiers (\eg, \citep{szegedy2013intriguing, goodfellow2014explaining, madry2017towards, carlini2017adversarial, athalye2018obfuscated}). 

In general adversarial attackers can be categorized into two groups based on the accessibility to the classifiers. {\em White-box} attackers (\eg, \citep{szegedy2013intriguing, goodfellow2014explaining, carlini2017adversarial}) have full knowledge of the classifiers such as network architectures as well as the trained weights. In this case, those attackers can directly calculate the perturbations that change the prediction using gradient ascent. {\em Black-box} attackers (\eg, \citep{ilyas2018black,su2019one}) have no right to access the classifiers, but they can observe the classification predictions given different inputs to estimate the perturbations. Black-box attackers are easier to apply on a classifier under privacy constraints but generally lead to lower success rate.

Adversarial training (\eg, \citep{goodfellow2014explaining, madry2017towards}) was introduced to defend against such attacks by augmenting the training data with adversarial samples. However, defense using adversarial training is resource-intensive since it requires large labeled datasets and retraining of the classifier.

To counter such attacks without changing existing classifiers, in this paper, we focus on adversarial {\em defenders}, which work as pre-processing modules to estimate the unattacked clean data to recover the prediction. The defenders can be categorized into either {\em gray-box} (or {\em oblivious} \citep{carlini2017adversarial}) defenders, where only the classifier but not the defender is accessible by the attackers, or {\em white-box} defenders, where the defender is also exposed to the attackers. State-of-the-art gray-box defenders use transformation models trained with labeled adversarial-clean image pairs (\eg,\citep{liao2018defense,akhtar2018defense}) to remove the perturbations. However, since such modules are differentiable, the attackers can directly regard the defender as part of the classifier and conduct gradient ascent under the white-box setting. Therefore previous attempts on white-box defenders mainly focus on the using of functions with {\em obfuscated gradients}, where either the functions are non-differentiable \citep{guo2018countering}, or the gradients are hard to compute due to very deep computation~\citep{samangouei2018defense}. However, those defenders that claim white-box robustness are recently broken by the Backward Pass Differentiable Approximation (BPDA) technique \citep{athalye2018obfuscated}, which approximates the derivatives by running the defender at the classifier forward pass and back-propagate the gradients using a differentiable approximation.

\begin{figure}[t]
    \centering
    \includegraphics[width=0.9\columnwidth]{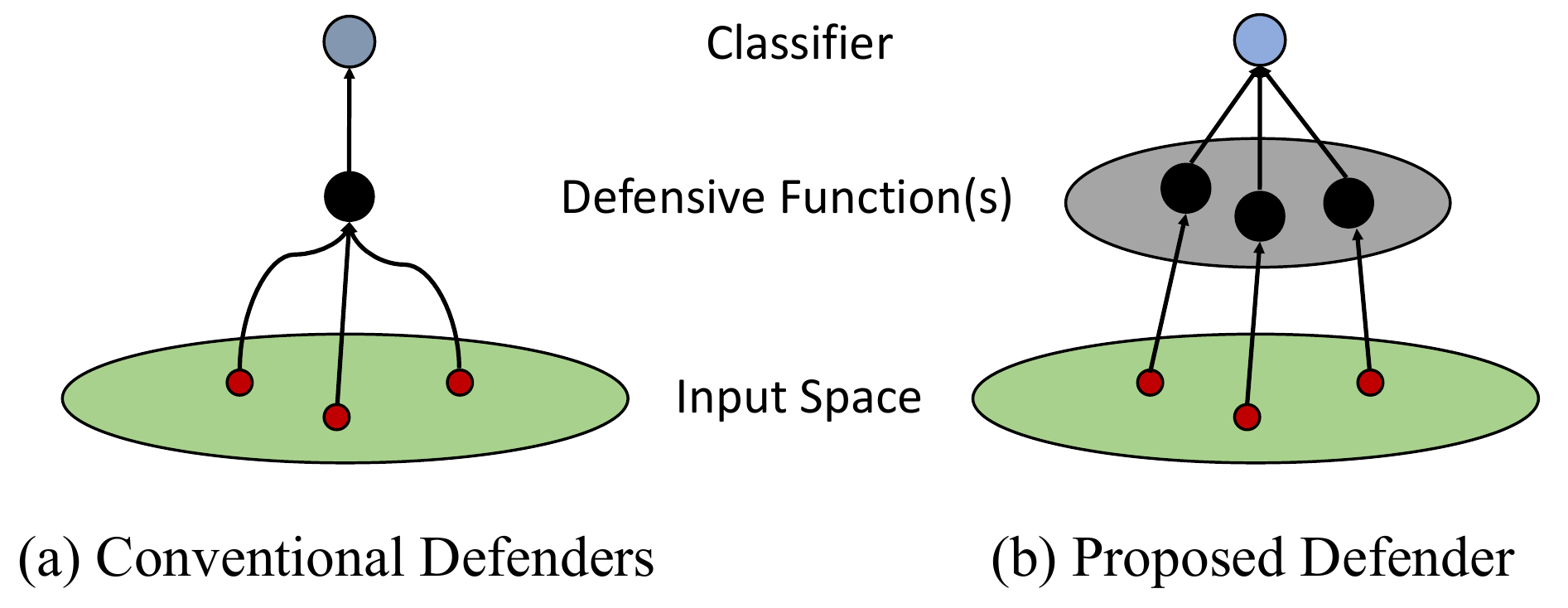}
    \vspace{-3mm}
    \caption{\footnotesize
    Illustration of the proposed {\em functional} defender. ({\bf a}) A conventional defender usually performs the defense with a fixed function. ({\bf b}) In a {\em functional} defender, every defensive function itself is a function of the input. Such a design significantly increases the difficulty of an adversarial attacker from gradient calculation/estimation.
    }\label{fig:overview}
    \vspace{-6mm}
\end{figure}

Since the invisibility assumption in gray-box defenders significantly increases the failure risk in practice, we therefore mainly focus on the {\em white-box} setting. That is, even the defender is exposed to the attackers, it can still prevent the attacker from not only direct derivative calculation but also gradient approximation. Besides, we hope such a defender does not require training and directly work at inference time. To this end, we propose viewing the defense from the perspective of a {\em functional}, a higher-order function that takes an individual adversarial sample as input and returns a new defender function exclusive for this sample (Fig.\ref{fig:overview}). Such a design makes the defender {\em input-dependent}, whose gradients are hard to be estimated without knowing the data prior (\eg, the prior distribution of natural images). Specifically, the functional is implemented to yield a parameterized neural network defender for each input, which is optimized by minimizing a {\em self-supervised} reconstruction loss using deep image prior~\citep{ulyanov2018deep} or denoising autoencoder~\citep{vincent2008extracting}. To further improve the robustness, we propose a novel Robust Iterative Data Estimation (\ride) algorithm that accelerates the convergence of the network output to the underlying clean data, which shares similar high-level concepts with momentum \citep{qian1999momentum}.

To summarize, our main contributions includes: (1) we are the first, to the best of our knowledge, to propose viewing a defender as a {\em functional} that returns an input-dependent function for every adversarial observation; (2) we propose a novel iterative self-supervised optimization algorithm, called \ride, as an effective realization of such a defender; (3) we demonstrate the significantly increased robustness of {\ride} against state-of-the-art BPDA white-box attackers, which achieves {98\%, 76\%, 43\%} Top-1 test accuracy on MNIST, CIFAR-10, and ImageNet datasets respectively.


%% file: related.tex
\section{Related Work}
\bfsection{Adversarial Attack} 
The vulnerability of deep learning models to small perturbations was first demonstrated by \cite{szegedy2013intriguing}. Later, \cite{goodfellow2014explaining} proposed an attacker of Fast Gradient Sign Method (FGSM) that utilizes the gradient sign to generate adversarial perturbations with only one-step update on the original image to fool classifiers. To improve this method, several iterative algorithms such as Basic Iterative Method (BIM) \citep{kurakin2016adversarial} and the Projected Gradient Descent (PGD) attacker \citep{madry2017towards} were proposed as well. \cite{moosavi2016deepfool} proposed DeepFool to find the minimal perturbation that changes the prediction. \cite{athalye2017synthesizing} proposed Expectation over Transformation (EOT) to attack against a random distribution of transformations. \cite{moosavi2017universal} showed the existence of universal adversarial perturbations that can fool a classifier with a single perturbation image. Such works above follow white-box attack, in contrast to black-box attackers (\eg, \citep{ilyas2018black, su2019one}). We refer readers to a nice survey by \cite{akhtar2018threat} for more details in this direction. 

In particular, \cite{athalye2018obfuscated} proposed Backward Pass Differentiable Approximation (BPDA) based on the fact that a defender $g$ always tries to recover the clean sample $\mathbf{x}$ from an adversarial sample $\Tilde{\mathbf{x}}$. Therefore, one can approximate the gradient of a defender \wrt the input using an approximation with an identity matrix. 

\bfsection{Adversarial Defense} 
Besides adversarial training (\eg, \citep{goodfellow2014explaining, madry2017towards, kannan2018adversarial, sinha2019harnessing}), \cite{xie2019feature} recently introduced non-local means into the classifiers as feature denoising blocks. More evaluations of the robustness of classifiers are conducted by \cite{hendrycks2018benchmarking}. For defenders that do not require changing of the classifier, \cite{guo2018countering} showed the defense using input transformations such as bit-depth reduction, total variance minimization (TVM), and image quilting while \cite{prakash2018deflecting} introduced pixel deflection. CNN-based denoisers (\eg, \citep{liao2018defense}) or rectifiers (\eg, \citep{akhtar2018defense}) can be trained with clean-adversarial image pairs to remove adversarial noise. \cite{samangouei2018defense} proposed Defense-GAN that uses a generator trained to model the distribution of clean images, and projects the adversarial sample into the space of the generator before classifying them. However, successful training of a GAN on datasets like ImageNet is still extremely challenging.

Different from previous literature, we propose a {\em functional}-based defender that returns an {\em unique} defender function for every individual input sample, which is the key feature that makes our method much more robust to white-box attacks. 

\bfsection{Self-Supervised Learning} 
Self-supervised learning is a form of unsupervised learning where the data itself provides the supervision \citep{zisserman2018}. Representative works in this direction include numerous variants of autoencoders~\citep{tschannen2018recent}. Recently, \cite{ulyanov2018deep} proposed {\em deep image prior} (DIP) demonstrating that a neural network can be used as a handcrafted prior with excellent results in inverse problems such as denoising, super-resolution, and inpainting. \cite{shocher2018zero} proposed {\em deep internal learning} (DIL) with similar observations. In this paper, we explore the applicability of those self-supervision algorithms as the implementation of the functional-based defenders.

%% file: motivation.tex
\section{Preliminaries}

\input{figure1.tex}

Generally adversarial attacks can be expressed as the following optimization problem (\eg,\citep{Gabriel2019first}):
\begin{align}\label{eqn:attack}
    \max_{\|\Delta\mathbf{x}\|\leq\epsilon} \ell\left(f(\mathbf{x}+\Delta\mathbf{x}), f(\mathbf{x})\right), 
\end{align}
where $\Delta\mathbf{x}$ denotes the perturbation for input $\mathbf{x}$, $f$ denotes the classifier, $\epsilon\geq0$ denotes a predefined (small) constant, $\ell$ denotes a proper loss function, and $\|\cdot\|$ denotes a norm operator. As we see, the attacker is essentially seeking for the adversarial sample $\mathbf{x}+\Delta\mathbf{x}$ that is {\em locally} around $\mathbf{x}$ but can change the prediction of the classifier as much as possible.

As discussed before, white-box attackers have full knowledge of the classifiers and can utilize the derivatives of the classifiers \wrt the inputs, {\em i.e.}, $\frac{\partial f}{\partial \mathbf{x}}$ in Eq. \ref{eqn:attack}, to determine the perturbation $\Delta\mathbf{x}$ using back propagation. On the contrary, learning a defender can be formulated as a minimization optimization problem as follows:
\begin{align}
    \min_{g\in\mathcal{G}} \ell\left(f(g(\mathbf{x}+\Delta\mathbf{x})), f(\mathbf{x})\right),
\end{align}
where $g\in\mathcal{G}$ denotes a defender function that tries to recover the original sample to minimize the prediction loss. In the literature such defenders are often assumed to be {\em invisible} to the attackers, which leads to {\em gray-box} defenders. clearly, if a fixed differentiable $g$ was known, an attacker could take $\tilde{f}=f\circ g$ as a new classifier to generate adversarial samples using $\frac{\partial \tilde{f}}{\partial \mathbf{x}}$. For $g$ with {\em obfuscated gradients}, $\frac{\partial {g}}{\partial \mathbf{x}}\approx \mathbf{I}$ is shown to be an effective approximation by the BPDA attacker. 


\bfsection{Adversarial Defenders as Functionals}
As we discussed above, any fixed-function defender can hardly survive white-box attacks. This motivates us to rethink the solution from the perspective of {\em funtional}, a higher-order function that takes functions as its argument to represent a function space (\ie, a space made of functions where each function can be thought of as a point). Conceptually, a {\em functional} defender is equivalent to an ensemble of different conventional function-based defenders. 
Now consider a functional $g(\mathbf{x}; \omega(\mathbf{x}))$ as the defender, whose input $\omega:\mathcal{X}\rightarrow\mathcal{W}$ is a (hidden) function over the a single input $\mathbf{x}\in\mathcal{X}$. In order to attack defender $g$, the derivative $\frac{\partial g}{\partial\mathbf{x}}$ can be written as follows:
\begin{align}\label{eqn:input-dependency}
    \frac{\partial g}{\partial\mathbf{x}} = \left[\mathbf{I}, \nabla\omega(\mathbf{x})\right]\nabla g(\mathbf{x}; \omega(\mathbf{x})),
\end{align}
where $\nabla$ denotes the gradient operator, $[\cdot, \cdot]$ denotes the matrix concatenation operator, and $\mathbf{I}$ denotes the identity matrix. With full access to $f$ and $g$, one may obtain $\nabla g$. However, the calculation of $\nabla\omega$ for hidden function $\omega$ without any prior knowledge at all will be challenging for the attack, in return leading to a significant increase in the success rate of defense.

\begin{figure}[t]
	\begin{minipage}[b]{0.48\columnwidth}
		\begin{center}
			\centerline{\includegraphics[width=.65\columnwidth]{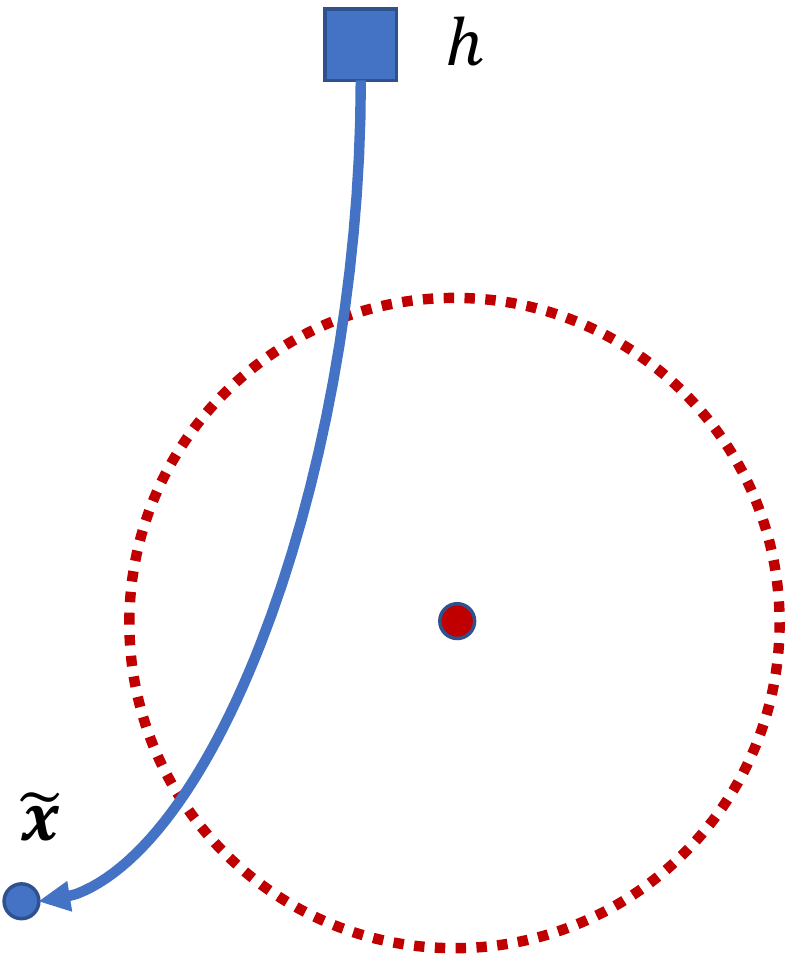}}
			\centerline{\footnotesize (a) DIP}
		\end{center}
	\end{minipage}
	\begin{minipage}[b]{0.48\columnwidth}
		\begin{center}
			\centerline{\includegraphics[width=.75\columnwidth]{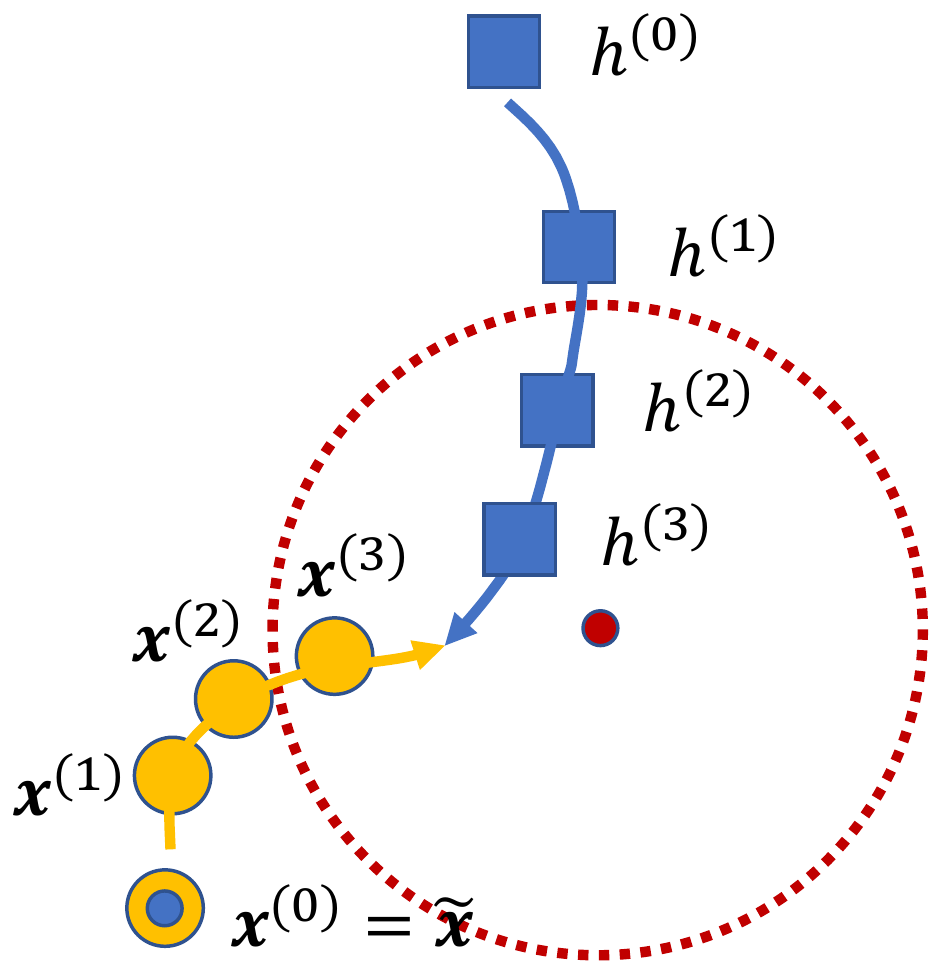}}
			\centerline{\footnotesize (b) Proposed {\ride}}
		\end{center}
	\end{minipage}	
    \vspace{-3mm}
	\caption{\footnotesize Hypothesis of learning trajectories (\ie the arrow curves) in the input space for {\bf (a)} DIP and {\bf (b)} our proposed {\ride} algorithm. Here the red dot denotes the unknown clean input, while the red dotted circle denotes the region where all the samples share the same prediction. {\ride} iteratively changes the estimation by convex interpolation to prevent overfitting and improve prediction recovery.} 
	\label{fig:path}
    \vspace{-5mm}
\end{figure}

\bfsection{Self-Supervision}
To realize the {\em functional} defenders, we refer to recent works in self-supervised learning that works on each individual sample without seeing external data. Specifically, DIP manages to estimate the true image by fitting a corrupted image $\tilde{\mathbf{x}}$, using a convolutional neural network (CNN) $h$, with randomly initialized weights $\mathbf{w}$, and an input consisting of random noise $\mathbf{z}$. Mathematically DIP tries to solve the following optimization problem:
\begin{align}\label{eqn:dip}
    \min_{\mathbf{w}\in\mathcal{W}}\ell_h\left(\tilde{\mathbf{x}}, h(\mathbf{z};\mathbf{w}, \theta)\right),
\end{align}
where $\ell_h$ denotes the loss function for optimizing the network, and $\theta$ denotes the network hyper-parameters during learning.
In the end, the functional $h$ returns a defender parameterized by $\mathbf{w}=\omega(\tilde{\mathbf{x}})$, which becomes an {\em input-dependent hidden} function. To the best of our knowledge, however, they have never been explored for adversarial defense. To verify their potential usage, given an adversarial image, we feed the intermediate reconstructed outputs from DIP into a pretrained ResNet50 model \citep{he2016deep} to see whether the prediction can be correctly recovered (Fig.~\ref{fig:motivation}). As we see in Fig. \ref{fig:motivation}(d), the output indeed manages to recover the original prediction of the classifier before starting to overfit the adversarial image. Meanwhile, we observe in Fig. \ref{fig:motivation}(e) that when the prediction is recovered, the distances of the output to the clean image ({\em dist2gt}) are consistently smaller than the distances to the adversarial image ({\em dist2adv}). This indicates that during optimization, the reconstructed image approaches the original image first, and then gradually diverges towards the adversarial image, as illustrated in Fig.~\ref{fig:path}(a). Therefore, how to effectively avoid such an overfitting problem in self-supervised learning becomes the key challenge of the defense. Later on we show two regularization techniques that significantly improve the performance of DIP as a defender.


%% file: figure1.tex
\begin{figure*}[t]
    \centering
    \includegraphics[width=\textwidth]{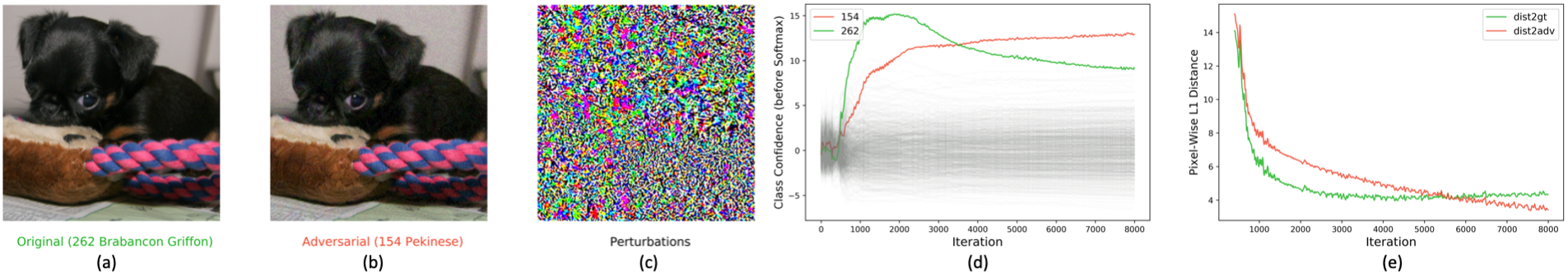}
    \vspace{-6mm}
    \caption{\footnotesize
    Characteristics of the reconstructed image using DIP when minimizing the reconstruction loss to the {\em adversarial image}. From left to right: {\bf (a)} clean image, {\bf (b)} adversarial image, {\bf (c)} (rescaled) adversarial perturbations, {\bf (d)} class confidence scores (before softmax) over iterations, and {\bf (e)} Manhattan distances of the reconstructed image to either clean (dist2gt) or adversarial image (dist2adv) over iterations. In (d), the green and red curves are for ground-truth and adversarial class labels, respectively, and the light gray curves are for the other classes. 
    }\label{fig:motivation}
    \vspace{-5mm}
\end{figure*}

%% file: method.tex
\section{White-Box Defense with Self-Supervision}

\bfsection{Problem Setup}\label{ssec:problem}
Let $(\mathbf{x},y)\in\mathcal{X}\times\mathcal{Y}$ be a clean data pair with a label $y$, and $\Tilde{\mathbf{x}}=\mathbf{x}+\Delta\mathbf{x}\in\tilde{\mathcal{X}}$ be an adversarial input. In the adversarial defense, we observe $(\Tilde{\mathbf{x}}, y)$ only, but not $\mathbf{x}$. 

Now given a pretrained classifier $f:\mathcal{X}\rightarrow\mathcal{Y}$ and a proper loss function $\ell:\mathcal{Y}\times\mathcal{Y}\rightarrow\mathbb{R}$, we aim to learn our defender $g:\tilde{\mathcal{X}}\times\mathcal{W}\rightarrow\mathcal{X}$ by optimizing the following problem:
\begin{align}\label{eqn:learning_prob}
    \min_{\omega\in\Omega} \mathbb{E}_{(\Tilde{\mathbf{x}},y)\in\tilde{\mathcal{X}}\times\mathcal{Y}}\Big[\ell\left(f(g(\Tilde{\mathbf{x}}, \omega(\Tilde{\mathbf{x}}))), y\right)\Big],
\end{align}
where $\mathbb{E}$ denotes the expectation operator, and  $\omega:\tilde{\mathcal{X}}\rightarrow\mathcal{W}$ denotes another (hidden) function. As discussed before, two fundamental differences between our method and the literature such as denoisers~\citep{liao2018defense} are: (1) no clean data is needed to learn our defender, and (2) our defender adaptively changes given each adversarial input. Such discrimination is derived from the nature of a functional.

\bfsection{Hyper-Parameter Learning as Realization}
From our analysis for Eq. \ref{eqn:input-dependency}, in order to design an effective white-box defender, function $\omega$ has to be a hidden function, namely, its explicit formula is unknown (otherwise, in the white-box defense the attacker can compute the gradient in Eq. \ref{eqn:input-dependency} as well). To this end, inspired by DIP we propose a novel self-supervised data estimation algorithm to embed the realization of $\omega$ and $g$ into the optimization of a neural network $h\left(\mathbf{z}; \mathbf{w}, \theta\right)$ by defining $\omega(\Tilde{\mathbf{x}})=\mathbf{w}$ that outputs different network parameters for every individual data point. 

With the help of $h\left(\mathbf{z}; \mathbf{w}, \theta\right)$ we can further rewrite Eq.~\ref{eqn:learning_prob} as a network hyper-parameter learning problem as follows:
\begin{align}\label{eqn:hyper-param}
    & \min_{\theta\in\Theta} \; \mathbb{E}_{(\Tilde{\mathbf{x}},y)\in\tilde{\mathcal{X}}\times\mathcal{Y}}\Big[\ell\left(f(\mathbf{x}_{est}), y\right)\Big],  \\
    & \mbox{s.t.} \hspace{.5mm} \mathbf{u}, \mathbf{w}\in\argmin_{\mathbf{u}', \mathbf{w}'}\mathcal{L}(\tilde{\mathbf{x}}, \mathbf{u}', \mathbf{w}', \theta), \mathbf{x}_{est} = \mathbb{E}_{\mathbf{z}\sim\mathcal{Z}}[h(\mathbf{z};\mathbf{w}, \theta)], \nonumber
\end{align}
where $\mathcal{L}$ denotes a reconstruction loss, $\mathcal{Z}$ denotes a data distribution, and $\mathbf{u}$ denotes an auxiliary variable that will be discussed in {\ride}. Here we consider the clean data estimator $\mathbf{x}_{est}$ as the mean of the network outputs over the entire distribution given the learned parameters, and define the output of our defender as $g(\Tilde{\mathbf{x}}, \omega(\Tilde{\mathbf{x}}))\equiv\mathbf{x}_{est}$.

\bfsection{Efficient Solver by Converting Learning to Tuning}
Eq. \ref{eqn:hyper-param} essentially defines a bilevel optimization problem \citep{colson2007overview}, and how to solve it effectively and efficiently, in general, is nontrivial. However, since the network hyper-parameters $\theta$ (\eg, number of iterations) often do not require continuity, to minimize $\ell$ we can 
\begin{enumerate}
    \item Predefine $\Theta$ by {\em discretizing} the hyper-parameter space;
    \item Evaluate each $\theta\in\Theta$ using the objective with learned $\mathbf{w}$'s;
    \item Choose the hyper-parameters with the minimum objective.
\end{enumerate}
This methodology simplifies the hyper-parameter learning problem in Eq. \ref{eqn:hyper-param} to a tuning problem. For instance, empirically, we tune the number of iterations as our stop criterion. Later on, we will discuss how to use self-supervision to design the lower-level objective $\mathcal{L}$ in Eq. \ref{eqn:hyper-param}.

\begin{algorithm}[t]\footnotesize
    \SetAlgoLined
    \SetKwInOut{Input}{Input}\SetKwInOut{Output}{Output}
    \Input{adversarial sample $\tilde{\mathbf{x}}$, initial weights $\mathbf{w}$, network $h$, loss $\ell_h$, Gaussian std $\boldsymbol\sigma$, number of iterations $K$, mini-batch size $n$, learning rate $\eta$}
    \Output{estimated true data $\mathbf{x}_{est}$}
    \BlankLine
    
    \For{$k=1,2,\cdots, K$}{
        
        Randomly draw $n$ samples $(\mathbf{s}, \mathbf{z})\sim\mathcal{N}(\tilde{\mathbf{x}}, \boldsymbol\sigma)\times\mathcal{N}(\tilde{\mathbf{x}}, \boldsymbol\sigma)$;


        $\mathbf{w}\leftarrow\mathbf{w} - \eta\cdot\frac{\partial\mathcal{L}_{DAE}}{\partial\mathbf{w}}$ \tcp*{Eq. \ref{eqn:rde}}
    }
    $\mathbf{x}_{est}=\mathbb{E}_{\mathbf{z}\sim\mathcal{N}(\tilde{\mathbf{x}}, \boldsymbol\sigma)}\left[h(\mathbf{z};\mathbf{w}, \theta)\right]$ \tcp*{mini-batch}
    
    \Return $\mathbf{x}_{est}$;
    \caption{Denosing Autoencoder (DAE)}\label{alg:dae}
\end{algorithm}

\begin{algorithm}[t]\footnotesize
    \SetAlgoLined
    \SetKwInOut{Input}{Input}\SetKwInOut{Output}{Output}
    \Input{adversarial sample $\tilde{\mathbf{x}}$, network $h$, least-square loss $\ell_h$, Gaussian std $\boldsymbol\sigma=0.5$, numbers of iterations and check-points $K=2000, T=5$, mini-batch size $n=1$, learning rate $\eta=10^{-4}$, interpolation parameter $\alpha=0.9$}
    \Output{estimated true data $\mathbf{x}_{est}$}
    \BlankLine
    
    Initialize the network weights $\mathbf{w}^{(0)}$, and $\mathbf{x}^{(0)}\leftarrow\tilde{\mathbf{x}}$;

    \For{$t=1,2,\cdots, T$}{
        $\mathbf{x}_{est}^{(t)}\leftarrow${\sc DAE}$\left(\mathbf{x}^{(t-1)}, \mathbf{w}^{(t-1)}, h, \ell_h, \boldsymbol\sigma, K, n, \eta\right)$;
        
        
        $\mathbf{x}^{(t)}\leftarrow\alpha\mathbf{x}^{(t-1)} + (1-\alpha)\mathbf{x}_{est}^{(t)}$;
    }
    \Return $\mathbf{x}_{est}^{(T)}$;
    \caption{Robust Iterative Data Estimation (\ride)}\label{alg:ride}
\end{algorithm}

\subsection{Self-Supervised Data Estimation}


\bfsection{Denosing Autoencoder for Robustness} 
In contrast to DIP that takes a fixed random noise image as input and the adversarial image as the target, we also propose employing the denoising autoencoder (DAE) \citep{vincent2008extracting} to improve the stability in data recovery using DIP. Different from the original goal of DAE which is to learn a low-dimensional encoding of the statistics of a dataset, here we apply DAE on every single input to model its {\em internal statistics} similar to DIL~\citep{shocher2018zero}. 
Considering further improvement on robustness over conventional DAE (see supplementary material for comparison), in our implementation we also add random Gaussian noise to the target, leading to the following lower-level objective:
\begin{align}\label{eqn:rde}
    & \mathcal{L}_{DAE}\left(\tilde{\mathbf{x}}, \emptyset, \mathbf{w}, \theta\right) \nonumber \\ & =\mathbb{E}_{(\mathbf{s},\mathbf{z})\sim\mathcal{N}(\tilde{\mathbf{x}}, \boldsymbol\sigma)\times\mathcal{N}(\tilde{\mathbf{x}}, \boldsymbol\sigma)}\Big[\ell_h\left(\mathbf{s}, h(\mathbf{z};\mathbf{w},\theta)\right)\Big],
\end{align}
where $\mathbf{s}$ and $\mathbf{z}$ denote the reconstruction target and the network input, respectively, and each $\mathcal{N}$ denotes a Gaussian distribution with mean $\tilde{\mathbf{x}}$ and std $\boldsymbol\sigma$. 
To solve Eq. \ref{eqn:rde}, we use mini-batch stochastic gradient descent (SGD) by randomly sampling the input-output pairs. We list our single-input {\sc DAE} algorithm in Alg. \ref{alg:dae} for reference.

\begin{figure}[t]
    \centering
    \includegraphics[width=\columnwidth]{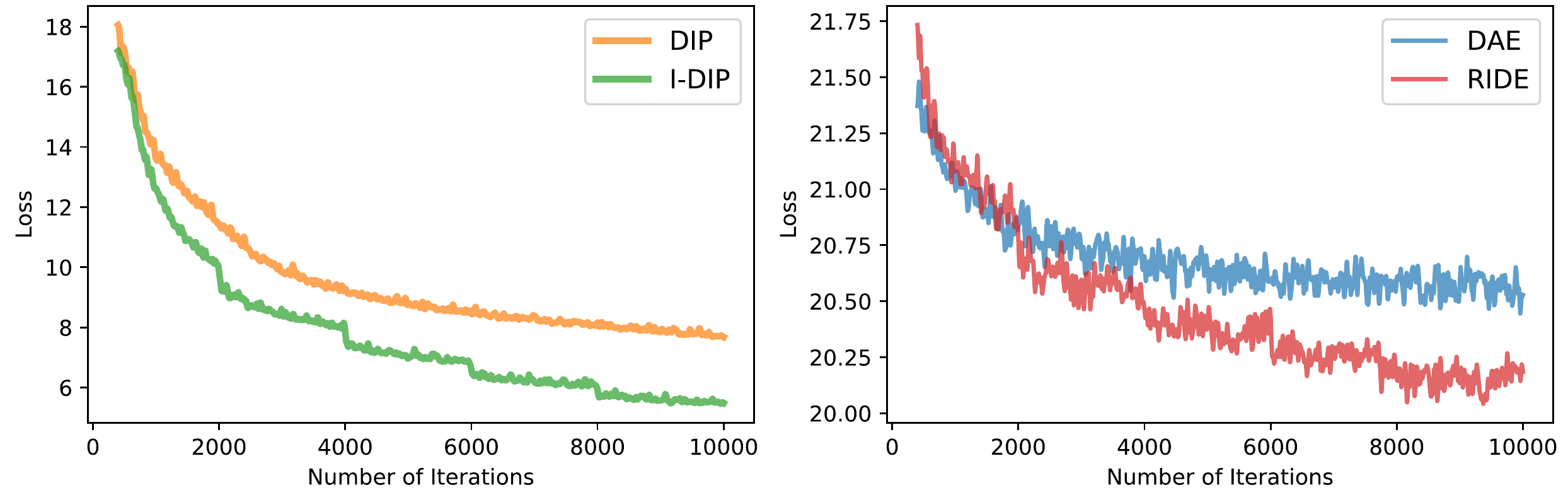}
    \vspace{-4mm}
    \caption{\footnotesize Illustration of convergence improvement of the proposed iterative optimization mechanism on ({\bf left}) DIP and ({\bf right}) DAE.
    }
    \label{fig:loss}
    \vspace{-3mm}
\end{figure}

\bfsection{Robust Iterative Data Estimation (\ride)}
DAE manages to improve the robustness in data estimation, but it converges slowly, as illustrated in Fig. \ref{fig:loss}(b), which may have a negative impact on the defense within a limited number of iterations. To remedy this, we propose {\ride}, a novel integration of {DAE} with iterative optimization that gradually estimates the true data (the auxiliary variable $\mathbf{u}=\mathbf{x}$ in Eq.~\ref{eqn:hyper-param}) as the target for fitting the network. As illustrated in Fig. \ref{fig:path}(b), the lower-level objective, $\mathcal{L}_{RIDE}$, can be formulated as follows: 
\begin{align}\label{eqn:ride}
    & \mathcal{L}_{RIDE}\left(\mathbf{x}^{(t-1)}, \mathbf{x}^{(t)}, \mathbf{w}^{(t)}, \theta\right) \nonumber \\
    & = \mathcal{L}_{DAE}\left(\mathbf{x}^{(t)}, \emptyset, \mathbf{w}^{(t)}, \theta\right) + \lambda d\left(\mathbf{x}^{(t)}, \mathbf{x}^{(t-1)}\right), \forall t,
\end{align}
where $\{\mathbf{x}^{(t)}\}$ denotes a sequence of the true data estimations with $\mathbf{x}^{(0)}=\Tilde{\mathbf{x}}$, $d$ denotes a distance measure as our regularizer, and $\lambda\geq0$ is a predefined constant. Comparing to Eq.~\ref{eqn:rde}, {\sc DAE} is a special case of {\sc Ride} with $\mathbf{x}^{(t)}=\tilde{\mathbf{x}}, \forall t$. In particular, if both $\ell_h$ and $d$ in Eq. \ref{eqn:ride} are mean squared error, we then have a close-form solution for updating $\mathbf{x}^{(t)}$ as follows:
\begin{align}\label{eqn:x_t}
    \mathbf{x}^{(t)} = \alpha\mathbf{x}^{(t-1)} + (1-\alpha)\mathbf{x}_{est}^{(t)}, \forall t,
\end{align}
where $\alpha = \frac{\lambda}{1+\lambda}$ and $\mathbf{x}_{est}^{(t)} = \mathbb{E}_{\mathbf{z}\sim\mathcal{N}(\mathbf{x}^{(t-1)}, \boldsymbol\sigma)}[h(\mathbf{z}; \mathbf{w}^{(t)}, \theta)]$. Such update rule appears to be widely used in deep learning optimizers as momentum (\eg, Adam \citep{kingma2014adam}) that helps accelerate gradients vectors in the right directions, thus leading to faster converging. Along with {\sc DAE}, we list our {\ride} algorithm in Alg.\ref{alg:ride}, where all the numbers denote the default hyper-parameter values in the defender\footnote{Theoretically $\mathbf{x}^{(T)}$ and $\mathbf{x}_{est}^{(T)}$ should merge when $T\rightarrow\infty$. Our empirical studies on ImageNet show that $\mathbf{x}^{(T)}$ is closer to $\tilde{\mathbf{x}}$ and performs worse than $\mathbf{x}_{est}^{(T)}$ by $\sim15\%$ in terms of test accuracy.}.

Note that such an iterative mechanism can also be integrated into DIP, leading to the {\em Iterative DIP} (I-DIP) algorithm. We illustrate the effects of our iterative mechanism in Fig. \ref{fig:loss} that clearly shows the convergence acceleration for both DIP and DAE in reconstruction.

\begin{thm}[Convergence of \ride]\label{thm:ride}
Suppose that $\forall t$, $$\mathcal{L}_{DAE}\left(\mathbf{x}^{(t-1)}, \emptyset, \mathbf{w}^{(t)}, \theta\right)\leq\mathcal{L}_{DAE}\left(\mathbf{x}^{(t-1)}, \emptyset, \mathbf{w}^{(t-1)}, \theta\right)$$ holds with sufficiently large number of iterations. Then {\ride} in Alg. \ref{alg:ride} is guaranteed to converge locally with $t\rightarrow\infty$. 
\end{thm}
\noindent
Please refer to the supplementary material for the proof.

%% file: experiment.tex
\section{Ablation Study on {\ride}}

\begin{figure*}[t]
    \centering
    \includegraphics[width=\textwidth]{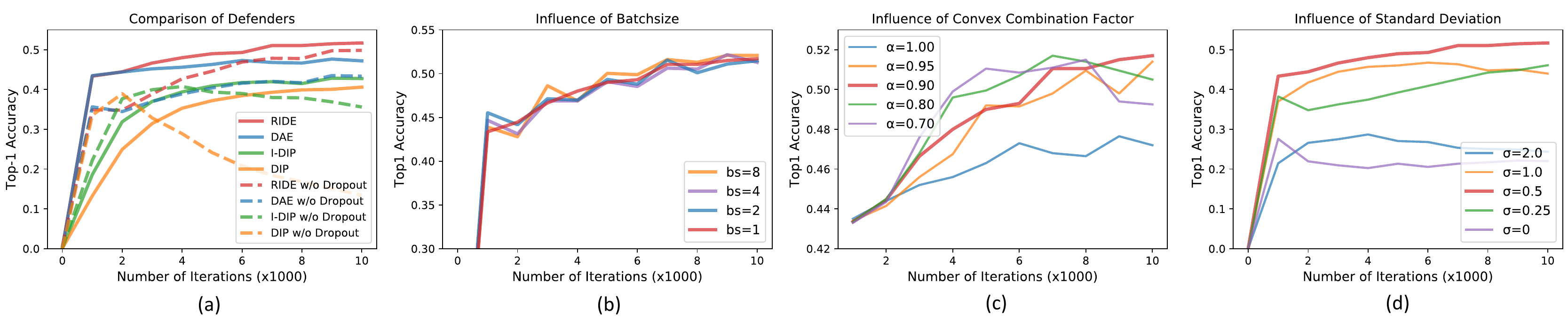}
    \vspace{-7mm}
    \caption{\footnotesize
    Ablation study on {\ride} in Alg. \ref{alg:ride} by comparing the effects of: 
    {\bf (a)} dropout, DAE, and iterative optimization; {\bf (b)} batch size $n$; {\bf (c)} interpolation parameter $\alpha$; and {\bf (d)} Gaussian std $\boldsymbol\sigma$ of DAE.
    }\label{fig:ablation}
    \vspace{-2mm}
\end{figure*}

\input{figure3.tex}

\bfsection{Implementation of {\ride}} 
By default we use an 8-layer fully convolutional network (FCN) similar to the one used in DIL \citep{shocher2018zero}. Except for the input and output layers, all the other layers use 32 filters with size $3\times3$. If using dropout, dropout layers with a ratio of 0.5 are added at the 4, 5 and 6-$th$ \texttt{conv} layers (check details in the supplementary material). 
No batch normalization layers are used. We initialize the model weights with random uniform distribution. 
The loss function is the  mean squared error for DIP, DAE and {\ride}. The model is optimized by the Adam optimizer \citep{kingma2014adam} with a learning rate $10^{-4}$ and weight decay $10^{-4}$. 
We fix the number of iterations $K=2,000$ and check-points $T=5$ for {\ride}. These parameters are fixed for all the following experiments in this paper without further fine-tuning.

\bfsection{Dataset}
We conduct experiments on the ImageNet \citep{imagenet_cvpr09} classification dataset. The dataset comprises 1.2 million training and 50,000 validation images, where each image has a label corresponding to one of the 1,000 classes. Following the setting of \cite{prakash2018deflecting}, we utilize a subset of the validation data by randomly sampling two images from each class, which consists of totally 2,000 images. Before being fed into the classifier and defenders, the images are center-cropped into $224\times 224$ pixels.

\bfsection{Classifier} 
By default, we use the pretrained ResNet50 model provided by PyTorch \citep{paszke2017automatic} to evaluate our defender\footnote{We also tested RIDE on VGG19~\citep{simonyan2014very} and Inception V3~\citep{szegedy2016rethinking} using the default hyper-parameter setting for ResNet50, which show $4.3\%$ and $8.2\%$ performance drop compared to the defense result of ResNet50. Please refer to the supplementary material for more details.}. The pretrained model achieves a top-1 classification accuracy of $76.15\%$ and $77.5\%$ on the validation set and our subset, respectively. For all the following experiments, the classifier stays in evaluation mode, and no gradient information is needed by our defenders.


\bfsection{Attacker} 
We employ the one-step FGSM attacker \citep{goodfellow2014explaining} with step size $0.015$. 

\subsection{Results}

\bfsection{Dropout \vs DAE \vs Iterative Optimization}
We compare the accuracy of the proposed defenders (DIP, I-DIP, DAE and \ride) against the number of iterations using the default hyper-parameters. As shown in Fig.~\ref{fig:ablation}(a), we observe that: 
\begin{itemize}
    \item Either dropout, or DAE, or iterative optimization can significantly help alleviate the overfitting in DIP, leading to better robustness in defense and better prediction recovery. 
    \item Among the three, DAE seems to be the most important for defense, and iterative optimization is the second.
    \item Our {\ride} algorithm, which involves all the three techniques, achieves the highest prediction accuracy.
\end{itemize}
We also show qualitative results in Fig.~\ref{fig:progress} to illustrate the reconstruction process of a single image using our proposed method. Note that the probability, $p$, of the ground-truth class increases with the increase of the number of iterations during the learning. Besides, the distance between the reconstructed image and the clean image ({\em dist2gt}) is consistently smaller than that between the reconstruction and the adversarial image ({\em dist2adv}), which further supports the motivation of the proposed algorithm as shown in Fig.~\ref{fig:motivation}(e). 

\bfsection{Reconstruction Model $h$}
We compare our FCN model with U-Net \citep{ronneberger2015u}, which features a symmetric encoder-decoder architecture with skip connections, by conducting the same experiments in Fig.~\ref{fig:ablation}(a). We observed a similar tendency of the curves, but the performance is $\sim 8\%$ lower for the U-Net model. Note that finding the optimal network architecture for reconstruction is out of the scope of this paper, and we will consider it in future work.

\bfsection{Mini-Batch Size $n$} 
We verify the performance of $n=\{1,2,4,8\}$ due to the limit of the GPU memory. Note that we still take a single image as input, while augmenting the input and target images by adding different Gaussian noise. As illustrated in Fig.~\ref{fig:ablation}(b), the performance using $n=1$ is comparable with the others. Considering that both time and memory consumption increase linearly with batch size, we fix the mini-batch size as 1 for all the following experiments.

\bfsection{Interpolation Parameter $\alpha$} 
As illustrated in Fig. \ref{fig:ablation}(c) we verify the performance of $\alpha=\{0.7, 0.8, 0.9, 0.95, 1\}$. Note that when $\alpha=1$, {\ride} becomes the DAE defender. As we see in Fig. \ref{fig:ablation}(c), $\alpha=0.9$ achieves the highest accuracy, while $\alpha=0.95$ performs marginally worse. $\alpha=0.8$ and $0.7$ reach peak performance at earlier stages, and we hypothesize that the estimator $\mathbf{x}^{(t)}$ becomes too far away from the unobserved sample, which may be regarded as an underfit problem.    


\bfsection{Gaussian Standard Deviation $\boldsymbol\sigma=\sigma\times\mathbf{I}, \sigma\geq0$}
For simplicity, we represent $\boldsymbol\sigma$ as a scalar times an identity matrix. Fig.~\ref{fig:ablation}(d) shows the influence of standard deviation in {\ride}. Recall that we add random Gaussian noise to both input and target images to further improve the robustness. When $\sigma=0$, our model becomes a vanilla autoencoder-decoder. When $\sigma$ is too small, the model tends to learn an identity mapping, while when $\sigma$ is too large (\eg, 2), the model is hard to converge. By default, we set $\sigma=0.5$ as this setting achieves the best performance among what we tested.

\bfsection{Running Time and Memory Footprint} For each image in the ImageNet dataset with size of $224\times 244$ pixels, our {\ride} takes $\sim 140s$ to finish the optimization on an NVIDIA Titan X GPU with a memory footprint of $\sim1$ gigabyte.

\section{State-of-the-art Performance Comparison}

\subsection{Defense Against Gray-Box Attack}
Following the experimental setup in the ablation study, we first compare the defensive performance on ImageNet against gray-box attacks as listed in Table \ref{tab:graybox}, where the attack has full access to the classifier, but not the defender. We use the default parameters in {\ride}, and the default classifier ResNet50 with $77.5\%$ top-1 accuracy using clean images. For the attacks, we use the public code released by~\cite{guo2018countering}, and for TVM defense, we use the Python scikit-image package~\citep{scikit-image}.

\begin{table}[t]
\caption{\footnotesize Performance comparison of different defenders on ImageNet against gray-box attacks. We show the top-1 accuracy of median filtering \citep{guo2018countering}, total variance minimization (TVM) \citep{guo2018countering}, and our {\ride} against both single-step FGSM \citep{szegedy2013intriguing} and iterative PGD \citep{madry2017towards}.}
\label{tab:graybox}
\vspace{-2mm}
\adjustbox{width=\columnwidth}{
\centering
\begin{tabular}{c|ccccc}
 \toprule
 & No Attack & No Defender & Median & TVM & RIDE ({\bf Ours}) \\ 
 \midrule
 FGSM & \multirow{2}{*}{0.775} & 0.075 & 0.306 & 0.359 & {\bf 0.517}\\ 
 PGD  & & 0.000 & 0.251 & 0.301 & {\bf 0.476}\\
 \bottomrule
\end{tabular}}
\vspace{-6mm}
\end{table}

\bfsection{Median Filtering} 
We test different kernel sizes for the median filter, \ie $k\in\{3, 5, 7, 9, 11\}$. For a given kernel size $k$, the input image is zero-padded by $\floor*{\frac{k}{2}}$ along the boundaries to keep the input size unchanged (\ie, $224\times 224$). We report the best result under each attacker in Table \ref{tab:graybox}. Please refer to our supplementary material for more results.

\bfsection{TV Minimization} Total variance minimization (TVM) tries to find an image similar to the input image but with less total variance. The trade-off is controlled by the weight factor where a larger one results in an image with less total variance at the cost of less similarity. The application of TVM on adversarial defense was recently introduced by \cite{guo2018countering} and tested on ImageNet. There are two commonly used TVM algorithms called the split-Bregman and Chambolle. In Table~\ref{tab:graybox} we show the best results among the two. Please refer to our supplementary material for more results.

\bfsection{Results \& Discussion}
As we see in Table~\ref{tab:graybox}, our {\ride} defender significantly outperforms the rest by at least $15.8\%$ and $17.5\%$ under the attack of FGSM and PGD, respectively. 

Recall that our {\ride} is developed for the stronger white-box attack, it should also work for the gay-box attacks. As a demonstration, we compare with median filtering and TVM, which have achieved (near) state-of-the-art performance on several benchmark datasets without classifier re-training or external data for defense. Those methods share the same setting as {\ride}. We are aware that there exist some other methods that may perform better. For instance, the image quilting defense also proposed by \cite{guo2018countering}. However, it requires a collection of one million clean images as an ``image bank'', whose patches are used to replace the regions in the adversarial images at inference time. Since it does not follow the setting that no external data should be used, for a fair comparison, we do not directly compare with such defenders. However, the performance of {\ride} is still highly comparable to the accuracy reported by \cite{guo2018countering}.

\begin{table}[t]
\caption{\footnotesize Performance comparison with the state-of-the-art defenders against the BPDA attacker. Our {\ride} defender achieves the best performance on all the three datasets under stronger attacks (\ie, larger maximum distances). Except {\ride}, all the other numbers are cited from the results reported by \cite{athalye2018obfuscated} under similar experimental settings.}
\label{tab:state-of-the-art}
\vspace{-3mm}
\adjustbox{width=\columnwidth}{
\centering
\begin{tabular}{lllr}
 \toprule
 {\bf Defense} & Dataset & Distance & Accuracy \\
 \hline
 Samangouei \etal (2018) & MNIST & 0.005 ($\ell_2$) & $55\%$ \\
 {\ride} ({\bf Ours}) & MNIST & 0.1 ($\ell_\infty$) & $\mathbf{98\%}$ \\
 \hline
 Buckman \etal (2018) & CIFAR & 0.031 ($\ell_\infty$) & $0\%$ \\
 Dhillon \etal (2018) & CIFAR & 0.031 ($\ell_\infty$) & $0\%$ \\
 Ma \etal (2018) & CIFAR & 0.031 ($\ell_\infty$) & $5\%$ \\
 Song \etal (2018) & CIFAR & 0.031 ($\ell_\infty$) & $9\%$ \\
 Na \etal (2018) & CIFAR & 0.015 ($\ell_\infty$) & $15\%$ \\
 Madry \etal (2018) & CIFAR & 0.031 ($\ell_\infty$) & $47\%$ \\
 {\ride} ({\bf Ours}) & CIFAR & 0.050 ($\ell_\infty$) & $\mathbf{76\%}$ \\
 \hline
 Xie \etal (2018) & ImageNet & 0.031 ($\ell_\infty$) & $0\%$ \\
 Guo \etal (2018) & ImageNet & 0.005 ($\ell_2$) & $0\%$ \\
 {\ride} ({\bf Ours}) & ImageNet & 0.050 ($\ell_\infty$) & $\mathbf{43\%}$ \\
 \bottomrule
\end{tabular}}
\vspace{-5mm}
\end{table}

\subsection{Defense Against White-Box Attack}

\bfsection{Datasets}
Besides our randomly selected dataset from ImageNet, here we also test our {\ride} on MNIST and CIFAR-10 to demonstrate the broad applicability of our algorithm. Both datasets contain 50k training and 10k test images, divided into 10 classes. Images in MNIST are grayscale of size $28\times 28$, while CIFAR contains color images of size $32\times 32$.

\bfsection{Classifiers}
For ImageNet, we continue to use ResNet50 as before. For MNIST, we train a simple CNN model with two convolutional and two fully connected layers, which reports a test accuracy of $98.98\%$. For CIFAR-10, we train a ResNet18 model with a test accuracy of $94.78\%$. 

\bfsection{Attackers}
We employ the BPDA attacker \citep{athalye2018obfuscated} to verify our white-box defense performance because as far as we know, BPDA  is currently the only attacker that can utilize the information (\ie, the outputs) of the defenders to attack the input images iteratively as a wrapper where in the inner loop another attacker is utilized. 

Specifically, for the inner loop, we apply the iterative $\ell_\infty$ PGD attack \citep{madry2017towards} for 10 iterations. For ImageNet and CIFAR-10, we perform the iterative attack with a step size of $0.01$ and maximum perturbation of $0.05$ on images normalized to $(0,1)$. For MNIST, we perform a stronger attack with maximum $\ell_\infty$ of $0.1$ and step size of $0.02$.

\bfsection{Networks for {\ride}}
For ImageNet, we use the same network as the one in the ablation study. Since the images in both MNIST and CIFAR-10 are much smaller, here we propose another two 4-layer (\ie, $1\rightarrow16\rightarrow16\rightarrow16\rightarrow1$ with ReLU activations) and 5-layer (\ie, $3\rightarrow32\rightarrow32\rightarrow32\rightarrow32\rightarrow3$ with ReLU activations) FCNs for them, respectively. 

\begin{figure}[t]
    \centering
    \includegraphics[width=\columnwidth]{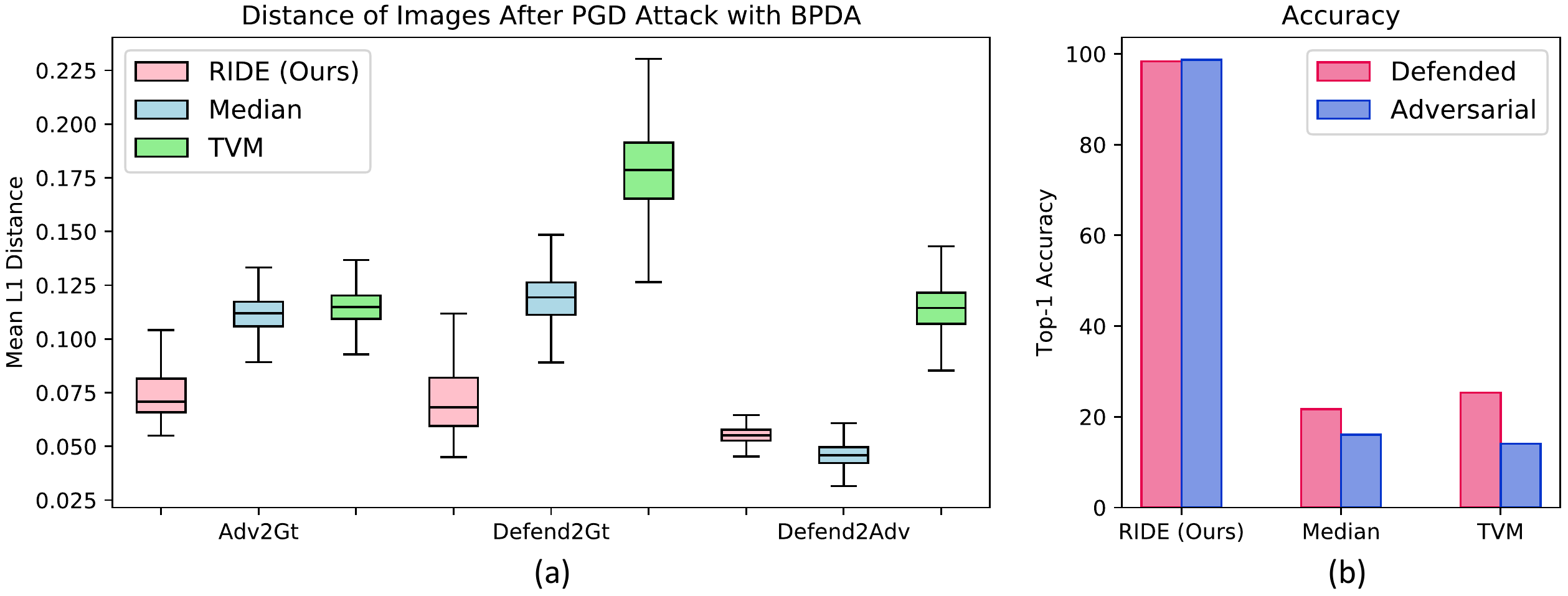}
    \vspace{-7mm}
    \caption{\footnotesize Analysis of defense against BPDA on MNIST: 
    {\bf (a)} Distances among adversarial, defended, and ground-truth images; {\bf (b)} Top-1 test accuracy of both the defended and adversarial images.
    }
    \label{fig:analysis}
    \vspace{-5mm}
\end{figure}

\bfsection{Results \& Discussion}
We list all the comparison results in Table \ref{tab:state-of-the-art}. The accuracy indicates that even under stronger attacks, our {\ride} still can significantly outperform the state-of-the-art defenders by large margins on all three datasets.

Recall that the BPDA attacker assumes that $g(\mathbf{x})\approx \mathbf{x}$ holds for a defender $g$ and estimate the gradient through $\nabla_{\mathbf{x}} f(g(\mathbf{x}))|_{\mathbf{x}=\Tilde{\mathbf{x}}} \approx \nabla f(\mathbf{x})|_{\mathbf{x}=g(\Tilde{\mathbf{x}})}$. To analyze why {\ride} can survive the attack, we compute the image distances on MNIST (Fig.~\ref{fig:analysis}(a)). After 10-iteration attacks, the distances of the defended images to the adversarial images (defend2adv) for {\ride} have a similar distribution as median filtering, both of which are much lower than TVM. This indicates that the assumption of $g(\mathbf{x}) \approx \mathbf{x}$ in BPDA is very likely to hold for {\ride}. However, the distances of the clean images to the adversarial (adv2gt) and defended images (defense2gt) using {\ride} are significantly lower than the other methods. 
This indicates that the gradients \wrt the input are hindered by {\ride}, making the adversarial images much ``cleaner''.

Besides, we feed both the adversarial and defended images after the attacks into the classifier. Fig.~\ref{fig:analysis}(b) indicates that for both kinds of images {\ride} achieves $\sim98\%$ accuracy while the other two perform much worse. Refer to Eq.~\ref{eqn:input-dependency}, we have:
\begin{align}
    & \nabla_{\mathbf{x}} f(g(\mathbf{x}; \omega(\mathbf{x}))) = \frac{\partial g}{\partial\mathbf{x}}\cdot\nabla f(g(\mathbf{x}; \omega(\mathbf{x}))) \nonumber \\
    & \hspace{10mm} = \left[\mathbf{I}, \nabla\omega(\mathbf{x})\right]\nabla g(\mathbf{x}; \omega(\mathbf{x}))\nabla f(g(\mathbf{x}; \omega(\mathbf{x}))).
\end{align}
With full access to both defender and classifier, the attacker can still compute $\nabla f(g(\mathbf{x}; \omega(\mathbf{x}))), \nabla g(\mathbf{x}; \omega(\mathbf{x}))$. However, calculating $\nabla\omega(\mathbf{x})$ can be extremely difficult because $\omega$ differs for each individual sample. So the explicit form of $\omega$ can not be estimated without knowing the prior distribution of the inputs. Such an input-dependent function breaks the assumption $\nabla_{\mathbf{x}} f(g(\mathbf{x}))|_{\mathbf{x}=\Tilde{\mathbf{x}}} \approx \nabla f(\mathbf{x})|_{\mathbf{x}=g(\Tilde{\mathbf{x}})}$ in BPDA, which is fundamental for the success of a {\em functional} defender.


%% file: figure3.tex
\begin{figure*}[t]
    \centering
    \includegraphics[width=\textwidth]{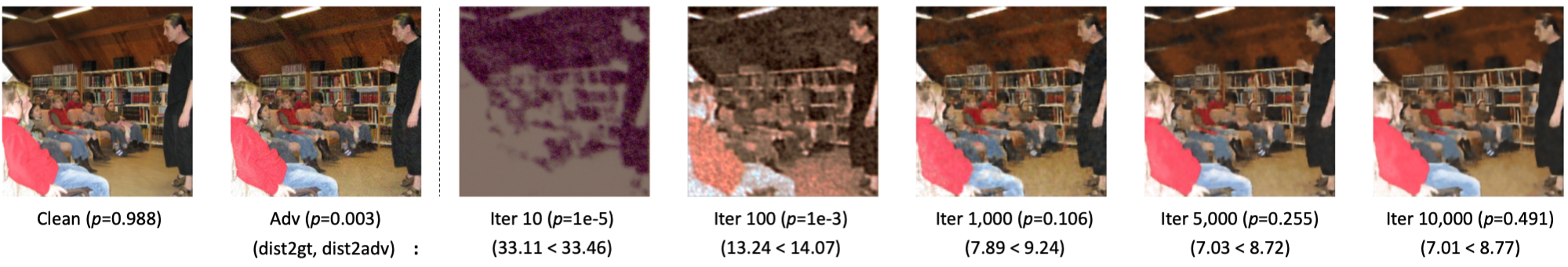}
    \vspace{-6mm}
    \caption{\footnotesize
    Visualization of the reconstruction process. From left to right: clean image, adversarial image, and reconstructed images via self-supervision at different iterations. The class probability of the correct class (624 {\em library}) are shown under each image, as well as the averaged Manhattan distances of the reconstructed image to the clean and adversarial images (dist2gt and dist2adv in $[0,255]$ space). 
    }\label{fig:progress}
    \vspace{-5mm}
\end{figure*}

%% file: appendix.tex
\newpage
\appendix
\section{Appendix}

\subsection{Proof of Theorem 1}

\begin{proof}
To prove this theorem, we consider the first three arguments in $\mathcal{L}_{RIDE}$ in Eq. \ref{eqn:ride} as variables, and try to minimize the loss by optimizing these variables one-by-one. To do so, based on the assumption in the theorem, Eq. \ref{eqn:ride} and Alg. \ref{alg:ride}, we have that $\forall t$,
\begin{align}
    & \mathcal{L}_{RIDE}\left(\mathbf{x}^{(t-1)}, \mathbf{x}^{(t-1)}, \mathbf{w}^{(t)}, \theta\right) \nonumber \\ 
    & \geq \mathcal{L}_{RIDE}\left(\mathbf{x}^{(t-1)}, \mathbf{x}^{(t)}, \mathbf{w}^{(t)}, \theta\right) \Longleftarrow Eq. \ref{eqn:x_t} \label{eqn:t1} \\
    & \geq \mathcal{L}_{RIDE}\left(\mathbf{x}^{(t)}, \mathbf{x}^{(t)}, \mathbf{w}^{(t)}, \theta\right) \Longleftarrow d(\mathbf{x}^{(t)}, \mathbf{x}^{(t-1)})=0 \label{eqn:t2} \\ 
    & \geq \mathcal{L}_{RIDE}\left(\mathbf{x}^{(t)}, \mathbf{x}^{(t)}, \mathbf{w}^{(t+1)}, \theta\right). \Longleftarrow Eq. \ref{eqn:rde}, \ref{eqn:ride} \label{eqn:t3}, assmp.
\end{align}
Eq. \ref{eqn:t1} optimizes the second variable based on Eq. \ref{eqn:x_t} while fixing the rest variables. Eq. \ref{eqn:t2} optimizes the first variable by making the distance term equal to 0 while fixing the rest. Eq.~\ref{eqn:t3} optimizes the third variable by training the network while fixing the rest. Moreover, since $\mathcal{L}_{RIDE}$ is lower-bounded by 0, we then can complete our proof.
\end{proof}

\subsection{Implementation Details}

\bfsection{Non-Maximum Suppression} To further prevent the reconstruction model from overfitting, we apply the non-maximum suppression at the convex combination step. That is, for current reconstructed and target images $\mathbf{x}_{est}^{(t)}$ and $\mathbf{x}^{(t)} \in \mathbb{R}^{3\times H\times W}$, we calculate the pixel-wise $\ell_1$ norm of the difference and get a set of distances at time step $t$.
\begin{equation}
S_d^{(t)} = \{d_{i} | d_{i} = \|{x}_{i}-{x}^\prime_{i}\|_1, x_{i} \in {\mathbf{x}}^{(t)}, x^\prime_{i}\in \mathbf{x}_{est}^{(t)}\}
\end{equation}
where $i$ is the index of the corresponding pixel. Because the optimization goal is to recover the image without overfitting to the adversarial target, the non-maximum suppression operation here is that for a given threshold $\tau$, the pixel pairs with $d_i < \tau$ are excluded in the convex combination steps and the pixels in the target image $\mathbf{x}^{(t)}$ are directly substituted by the corresponding pixels in the reconstructed image $\mathbf{x}_{est}^{(t)}$. The reason is that when the distance between the reconstructed image and target image are already close enough, substitute pixels in the adversarial image with pixels in the reconstructed image can decrease the tendency of overfitting but cause little influence on the learning of image content and texture. For all the experiments with iterative updating described in this paper, we just set the threshold $\tau = \text{median}\{S_d\}$, which is calculated adaptively on each individual image without substantial parameter searching. 

\begin{figure}[t]
    \centering
    \includegraphics[width=0.6\columnwidth]{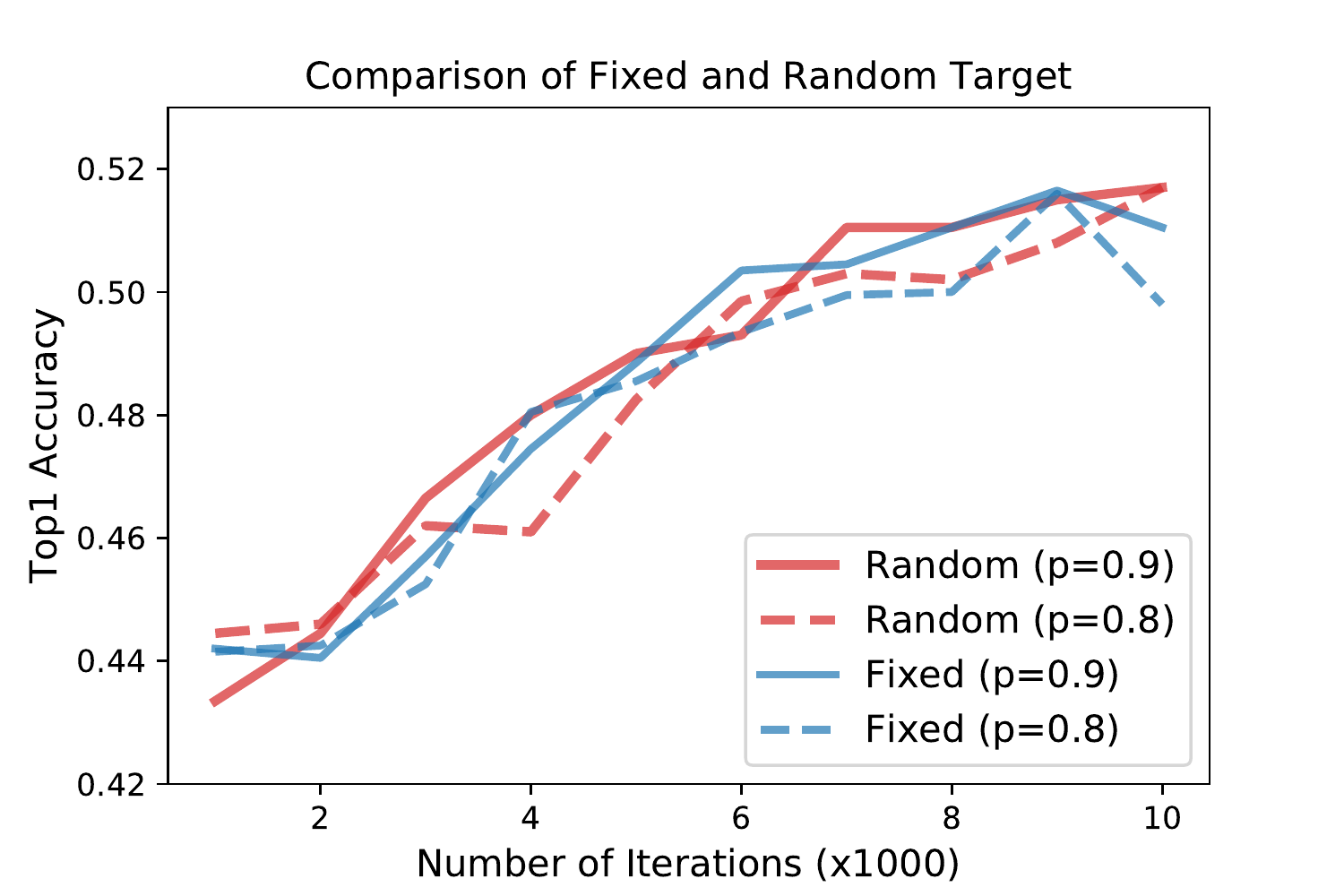}
    \vspace{-1mm}
    \caption{\footnotesize Comparison of the proposed {\ride} algorithm using fixed target image (blue curves) or target image with random Gaussian noise (red curves). We show the comparison with random loss masking ratio of $p=0.9$ and $p=0.8$.
    }
    \label{fig:append-random-target}
    \vspace{-3mm}
\end{figure}

\bfsection{Random Loss Masking} Another technique we apply to the self-supervised image reconstruction process is to randomly mask out part of the pixels in the loss calculation. Because the reconstruction network works on a single image, therefore no matter what are the noise patterns in the adversarial image, the model can easily overfit to it. Therefore we can randomly exclude $p$ portion of the pixels in loss calculation at each iteration. Although after updating the model for thousands of iterations by expectation every pixel in the target image should be directly supervised for several times, the random loss masking can still alleviate the risk of overfitting because every iteration the pixel value for some pixels are learned not based on its noisy version but the context in the corresponding field of view. We show experiments on {\ride} with fixed convex combination factor $\alpha=0.9$ using different random masking ratio $p$ in Fig.~\ref{fig:append-random-target}. Accordingly, for all the experiments in the paper, we fix the random masking ratio of $p=0.9$.

\bfsection{Model Architecture} Here we show the architectures of the reconstruction models used in our experiments (Table~\ref{tab:architecture}). The random dropout ratio for Dropout layers is 0.5.

\bfsection{Image Normalization} Pre-trained PyTorch classifiers (\eg, ResNet50, VGG19, and InceptionV3) suppose that an input image is normalized to have zero mean. Specifically, for our ablation study and gray-box defense experiments on ImageNet dataset, the adversarial images are pre-computed, and before feeding the images into the {\ride} defender, the pixel values of the images are normalized by means $\mu=(0.485, 0.456, 0.406)$ and standard deviation $\sigma=(0.229, 0.224, 0.225)$. Therefore there is no {\em sigmoid} activation for the reconstruction model in {\ride} for ablation study and gray-box defense experiments. However, for the white-box setting ({\em i.e.}, BPDA) the perturbations need to be calculated on-the-fly, therefore we add a {\em sigmoid} activation to the reconstruction model to make the values of the output images restricted to $(0,1)$, and make the normalization a differentiable layer at the beginning of the classifier.

We apply the same operation for CIFAR-10 and MNIST under the white-box defense setting. The mean and standard deviation for CIFAR-10 are the same as those used for ImageNet. For MNIST, which consists of gray-scale images, the mean is $0.1307$ while the standard deviation is $0.3081$.


\subsection{Additional Results}

\begin{table}[t]
\caption{\footnotesize Architecture of reconstruction models. All layers except the output layer (\texttt{convn}) are followed by a ReLU activation. There is a sigmoid activation following the output layer if the image range is $(0,1)$, and no activation for image normalized with $mean=0$.}\label{tab:architecture}
\adjustbox{width=\columnwidth}{
\centering
\begin{tabular}{ccccc}
 \toprule
 Dataset & Layer Name & Input Planes & Output Planes & Dropout \\ 
 \midrule
 \multirow{8}{*}{ImageNet} & \texttt{conv1} & 3 & 32 &  \\
  & \texttt{conv2} & 32 & 32 &  \\
  & \texttt{conv3} & 32 & 32 &  \\
  & \texttt{conv4} & 32 & 32 & \cmark \\
  & \texttt{conv5} & 32 & 32 & \cmark \\
  & \texttt{conv6} & 32 & 32 & \cmark \\
  & \texttt{conv7} & 32 & 32 &  \\
  & \texttt{convn} & 32 & 3 &  \\
 \midrule
 \multirow{4}{*}{MNIST} & \texttt{conv1} & 1 & 16 &  \\
 & \texttt{conv2} & 16 & 16 &  \cmark \\
 & \texttt{conv3} & 16 & 16 &  \cmark \\
 & \texttt{convn} & 16 & 1  &  \\
 \midrule
 \multirow{5}{*}{CIFAR-10} & \texttt{conv1} & 3 & 32 &  \\
 & \texttt{conv2} & 32 & 32 &  \cmark \\
 & \texttt{conv3} & 32 & 32 &  \cmark \\
 & \texttt{conv4} & 32 & 32 &  \\
 & \texttt{convn} & 32 & 3  &  \\
 \bottomrule
\end{tabular}}
\end{table}

\begin{table}[h]
\caption{\footnotesize Performance of median filtering against {\em gray-box} attacks. We show the top-1 accuracy of different $kernel\_size$ on ImageNet validation set. The highest value is reported in Table~\ref{tab:graybox}.}\label{tab:median}
\centering
\adjustbox{max width=0.7\columnwidth}{
\centering
\begin{tabular}{cccccc}
 \toprule
 Attack & 3 & 5 & 7 & 9 & 11 \\ 
 \midrule
 FGSM & 0.180 & 0.281 & {\bf 0.306} & 0.277 & 0.241\\
 PGD  & 0.005 & 0.128 & 0.238 & {\bf 0.251} & 0.228\\
 \bottomrule
\end{tabular}}
\end{table}

\begin{table}[h!]
\caption{\footnotesize Performance of total-variation denoising using split-Bregman optimization against {\em gray-box} attacks. We show the top-1 accuracy of different $denoising\_weight$ on ImageNet validation set. The highest value is reported in Table~\ref{tab:graybox}.}\label{tab:bregman}
\centering
\adjustbox{max width=0.7\columnwidth}{
\centering
\begin{tabular}{cccccc}
 \toprule
 Attack & 0.075 & 0.125 & 0.25 & 0.5 & 0.75 \\ 
 \midrule
 FGSM & 0.312 & 0.339 & {\bf 0.359} & 0.326 & 0.288 \\
 PGD  & 0.281 & {\bf 0.301} & 0.289 & 0.218 & 0.149 \\
 \bottomrule
\end{tabular}}
\end{table}

\begin{table}[h!]
\caption{\footnotesize Performance of Chambolle total-variation denoising against {\em gray-box} attacks. We show the top-1 accuracy of different $denoising\_weight$ on ImageNet validation set. The highest values under two attacks are lower than the Bregman method.}\label{tab:chambolle}
\centering
\adjustbox{max width=0.92\columnwidth}{
\centering
\begin{tabular}{cccccccc}
 \toprule
 Attack & 0.25 & 0.5 & 0.75 & 1.0 & 1.25 & 1.5 & 1.75 \\
 \midrule
 FGSM & 0.174 & 0.277 & 0.321 & {\bf 0.341} & 0.332 & 0.323 & 0.315\\
 PGD  & 0.017 & 0.138 & 0.234 & 0.267 & 0.279 & 0.283 & {\bf 0.285}\\
 \bottomrule
\end{tabular}}
\end{table}

\begin{figure}[t]
    \centering
    \includegraphics[width=\columnwidth]{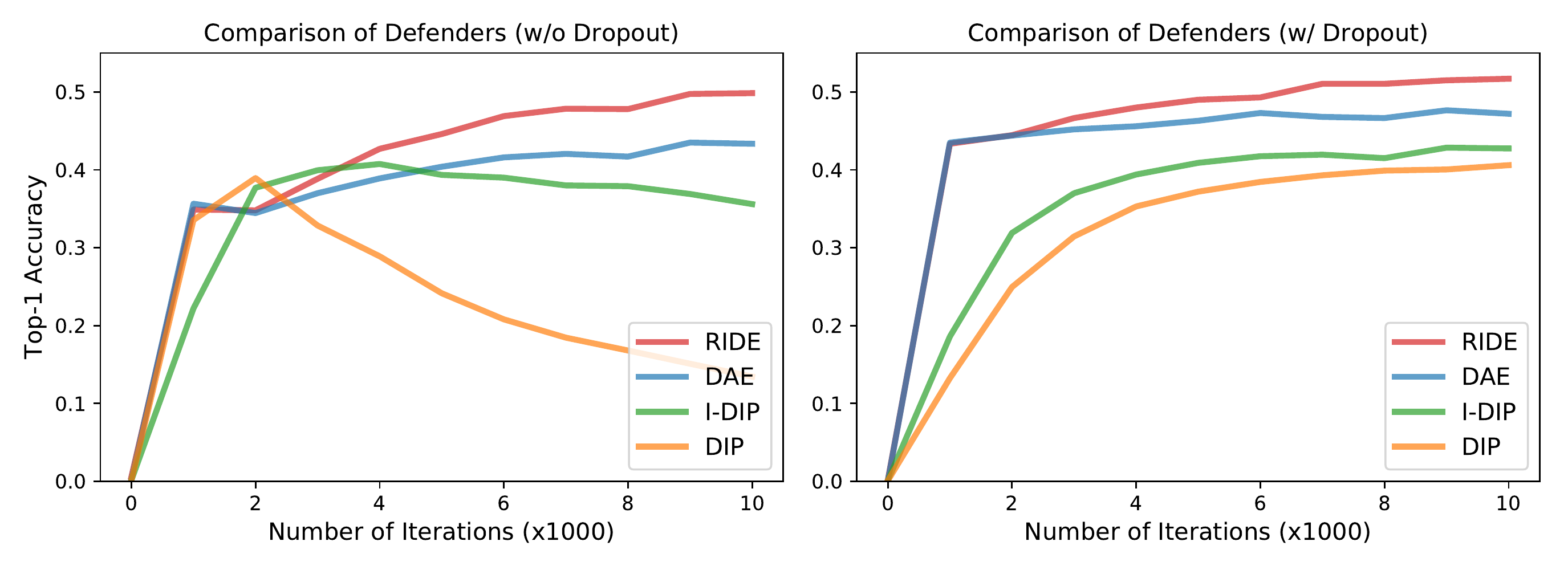}
    \vspace{-2mm}
    \caption{\footnotesize Comparison of different proposed defenders on classification performance {\bf (left)} without, or {\bf (right)} with dropout layers in the reconstruction model. This figure just separates the curves in Fig.~\ref{fig:ablation}(a) to clearly indicate the incremental performance improvement of different proposed techniques. 
    }
    \label{fig:append-dropout}
    \vspace{-3mm}
\end{figure}

\bfsection{Other Classifier \& Clean Images} To show the transferability of the proposed defense algorithm, we also tested the RIDE defender on VGG19~\citep{simonyan2014very} and Inception V3~\citep{szegedy2016rethinking} models. Using the optimal hyper-parameter settings for ResNet50, the RIDE defender show only $4.3\%$ and $8.2\%$ performance drop on VGG19 and Inception V3 models compared to the FGSM defense result reported on ResNet50 ({\em i.e.}, $51.7\%$). Not that the adversarial samples are calculated using the target classifiers instead of using the ones generated for ResNet50. Besides, when taking the clean images as inputs, the test accuracy is $61.75\%$ using the ResNet50 model.

\bfsection{Median Filtering} We show the results of using median filtering as defenders under gray-box setting on the ImageNet validation dataset in Table~\ref{tab:median}. For a given kernel size $k$ (an odd number), the input image is zero-padded by $\floor*{\frac{k}{2}}$ along the boundaries to keep the input size unchanged ($224\times 224$).

\bfsection{TV Minimization} Total variance minimization (TVM) tries to find an image similar to the input image but with less total variance. The trade-off is controlled by the weight factor $w$ where a larger $w$ results in an image with less total variance at the expense of less similarity. The application of TVM on adversarial defense was recently introduced by \cite{guo2018countering} and are tested on the ImageNet dataset. There are two commonly used TVM algorithms called the split-Bregman and Chambolle. Here we show the defense performance of the two algorithms in Table~\ref{tab:bregman} and~\ref{tab:chambolle}. The highest scores are reported in Table~\ref{tab:graybox}.

\subsection{Qualitative Results}

We show the qualitative comparisons on MNIST, CIFAR, and ImageNet datasets in the following figures.

\newpage
\begin{figure*}[t]
    \centering
    \includegraphics[width=0.8\textwidth]{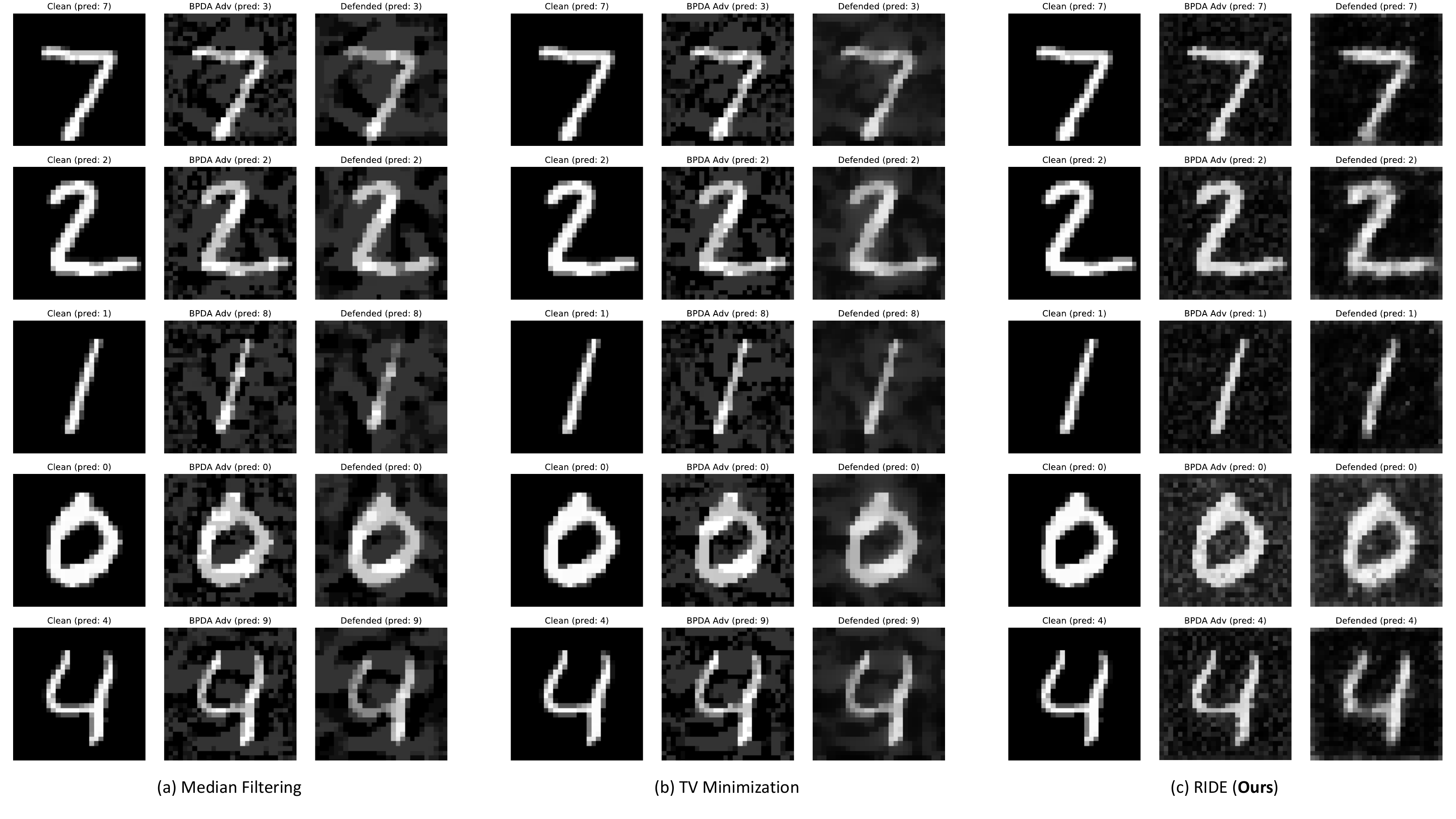}
    \vspace{-2mm}
    \caption{\footnotesize Qualitative results of defense against 10-iteration PGD attack (with BPDA) on MNIST dataset using median filtering {\bf (a)}, total-variance minimization {\bf (b)} and the proposed {\ride} algorithm {\bf (c)}. The predicted label of each image is shown on top of it.
    }
    \label{fig:append-mnist}
\end{figure*}

\newpage
\begin{figure*}[t]
    \centering
    \includegraphics[width=0.8\textwidth]{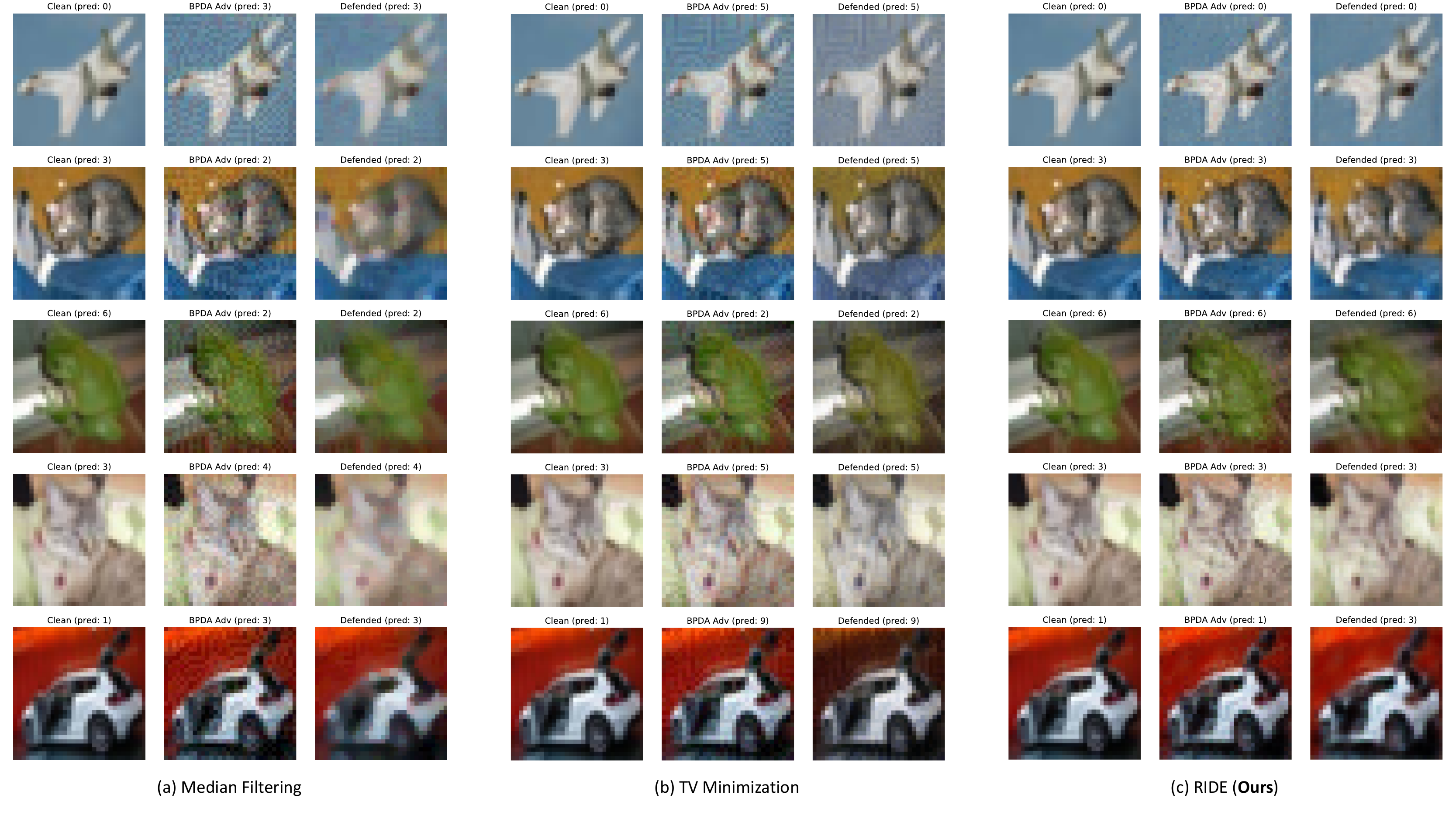}
    \vspace{-2mm}
    \caption{\footnotesize Qualitative results of defense against 10-iteration PGD attack (with BPDA) on CIFAR-10 dataset using median filtering {\bf (a)}, total-variance minimization {\bf (b)} and the proposed {\ride} algorithm {\bf (c)}. The predicted label of each image is shown on top of it. The last row of {\ride} denote a failure case.
    }
    \label{fig:append-cifar}
\end{figure*}

\newpage
\begin{figure*}[t]
	\begin{minipage}[b]{0.33\textwidth}
		\begin{center}
			\centerline{\includegraphics[width=\columnwidth]{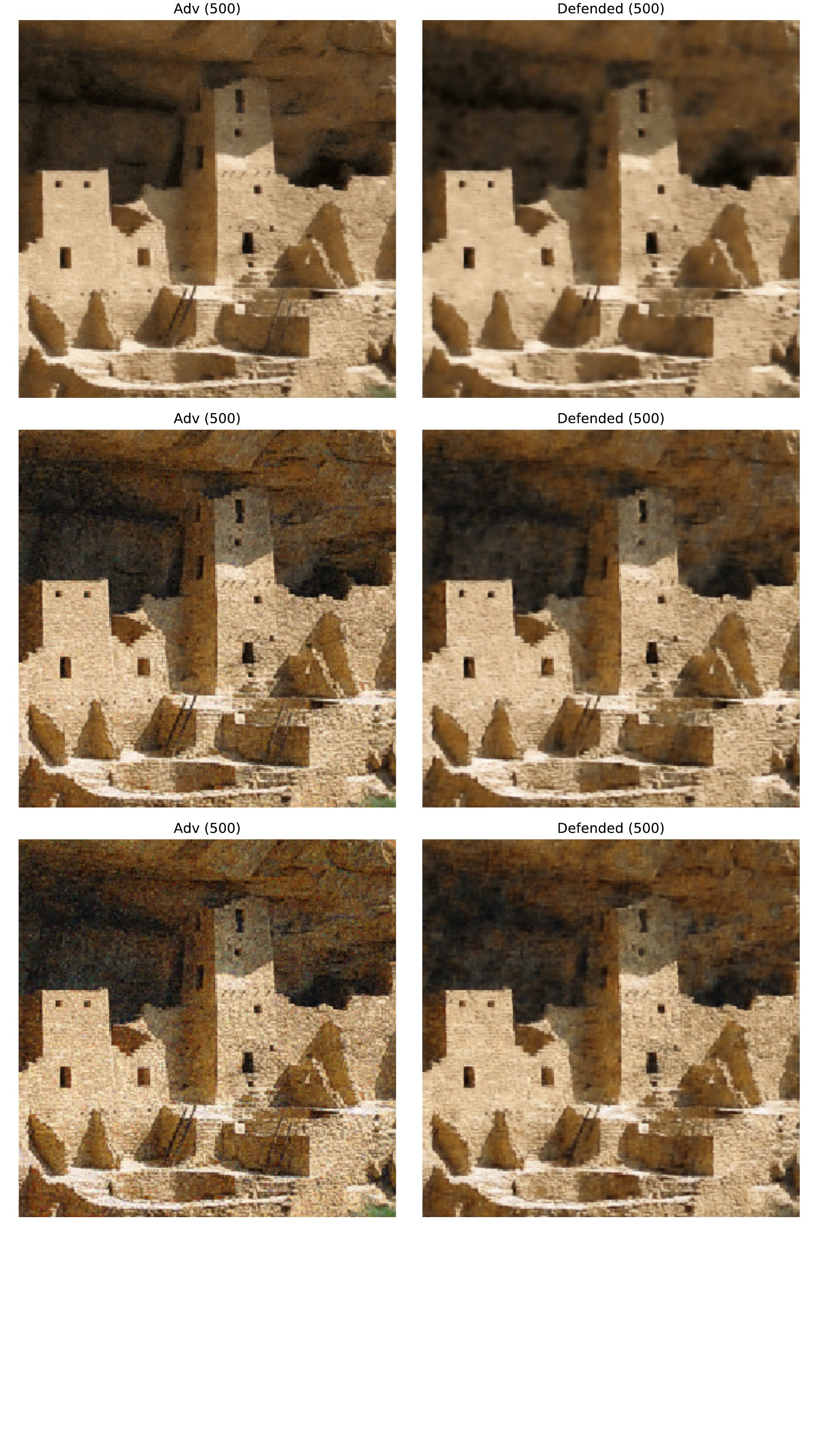}}
			\vspace{-15mm}
			\centerline{\footnotesize ({\bf a}) cliff dwelling (500)}
		\end{center}
	\end{minipage}
		\begin{minipage}[b]{0.33\textwidth}
		\begin{center}
			\centerline{\includegraphics[width=\columnwidth]{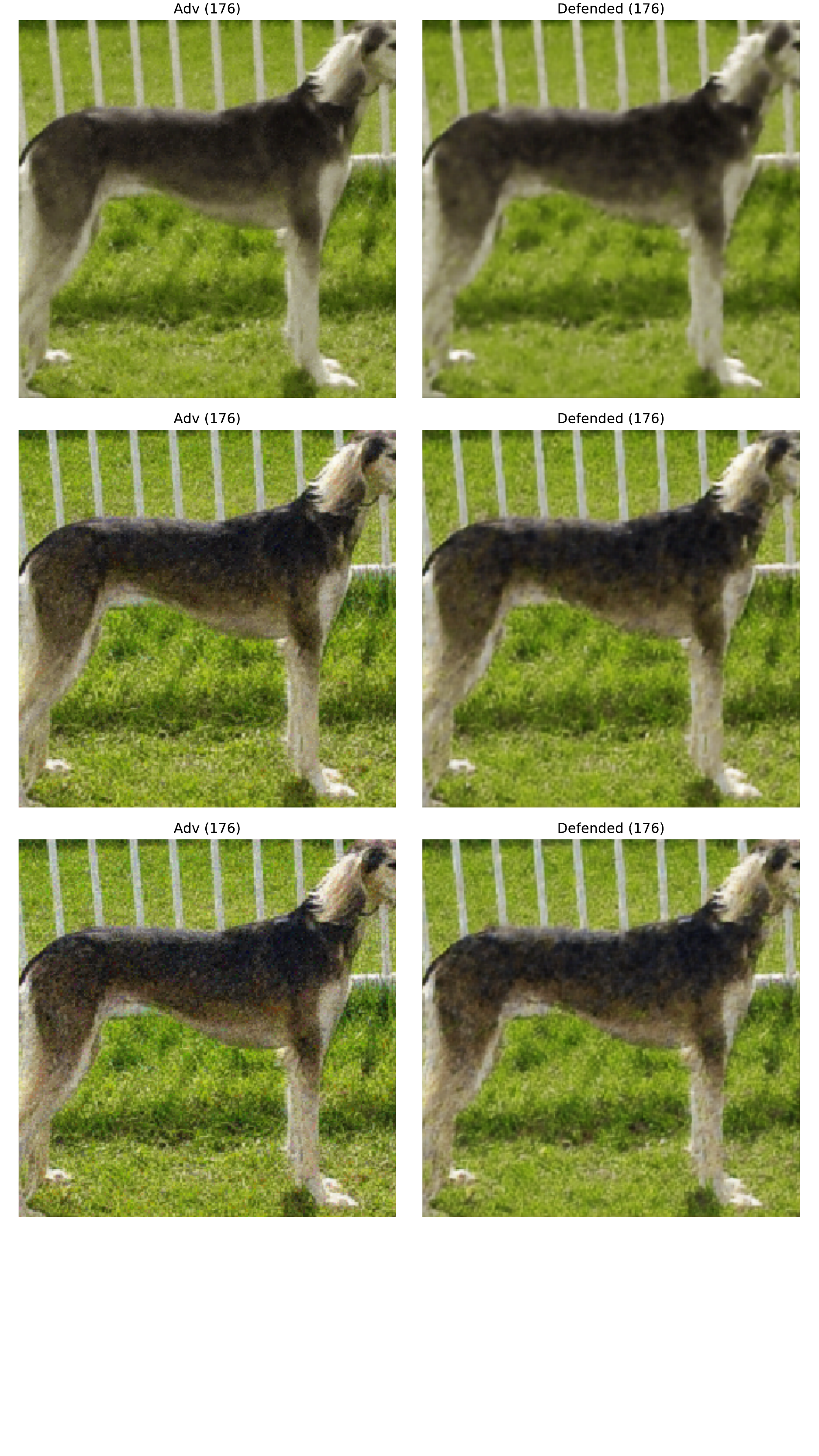}}
			\vspace{-15mm}
			\centerline{\footnotesize ({\bf b}) Saluki, gazelle hound (176)}
		\end{center}
	\end{minipage}	
		\begin{minipage}[b]{0.33\textwidth}
		\begin{center}
			\centerline{\includegraphics[width=\columnwidth]{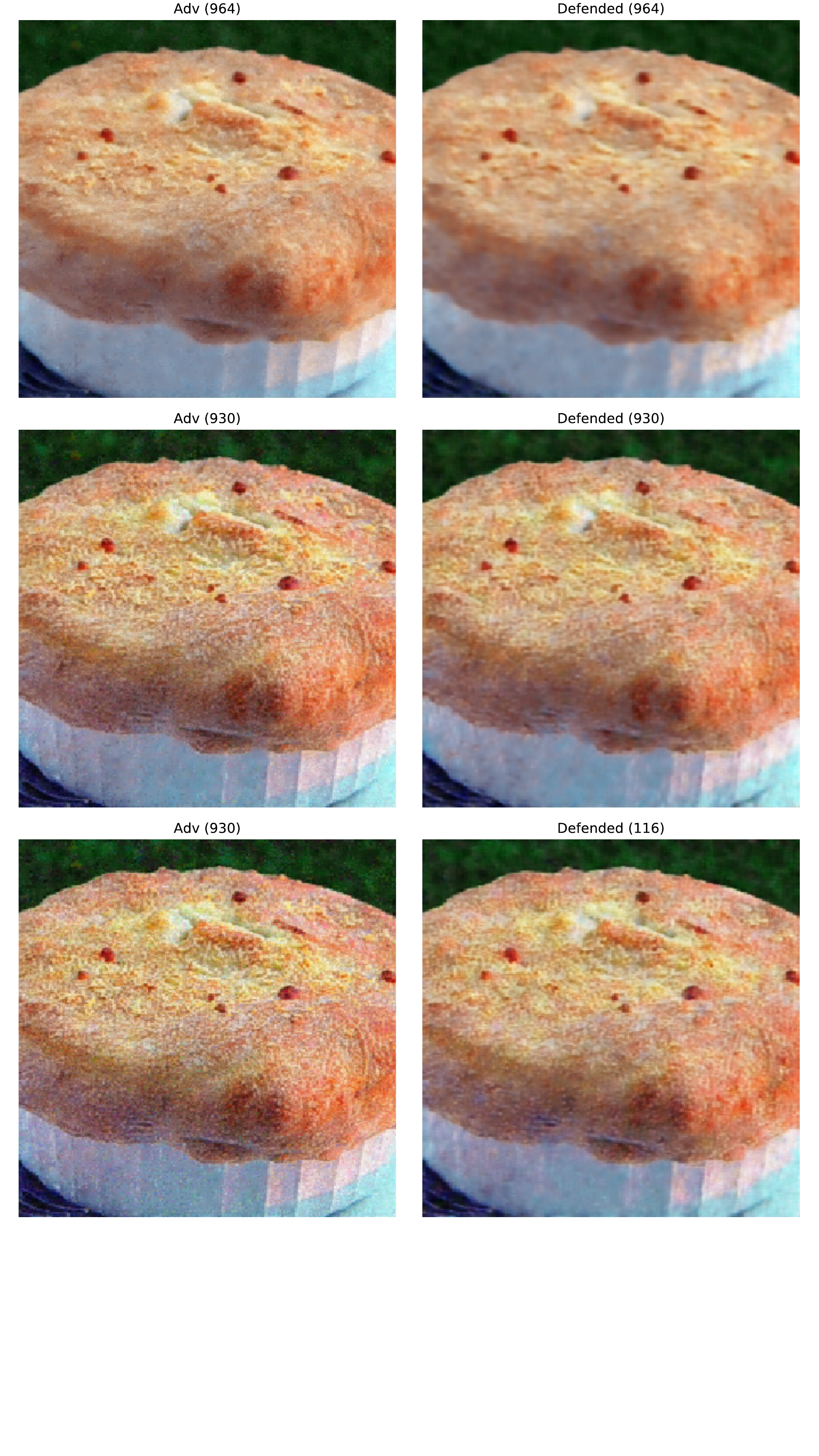}}
			\vspace{-15mm}
			\centerline{\footnotesize ({\bf c}) potpie (964)}
		\end{center}
	\end{minipage}	
    \vspace{-6mm}
	\caption{\footnotesize Qualitative results of defense against 10-iteration PGD attack (with BPDA) on ImageNet dataset, using the proposed {\ride} defender. Every two-column shows both the adversarial and defended images after 1,4 and 10 iterations of attack. The predicted label of each image is shown on top of it, where ({\bf c}) denotes a failure case.} 
	\label{fig:imagenet}
\end{figure*}